\PassOptionsToPackage{sort&compress}{natbib}
\documentclass[review]{elsarticle}

\usepackage{lineno,hyperref}
\modulolinenumbers[5]

\journal{Artificial Intelligence}

\usepackage{times}
\usepackage{graphicx}
\usepackage{latexsym}
\usepackage{csquotes}
\usepackage{color}
\usepackage{array}

\usepackage{geometry}
\usepackage[many]{tcolorbox}
\usepackage{lscape}

\usepackage{tikz}
\usetikzlibrary{calc}
\usetikzlibrary{decorations.pathmorphing}
\usetikzlibrary{decorations.pathreplacing}
\usepackage{hyperref}

\usepackage{xcolor}
\usepackage[framemethod=tikz]{mdframed}

\definecolor{cccolor}{rgb}{.67,.7,.67}

\newcommand{\denquote}[1]{\textquoteleft#1\textquoteright}

\begin{document}

\begin{titlepage}

\begin{center}
\vspace*{1cm}

\LARGE
\textbf{What Do We Want From Explainable Artificial Intelligence (XAI)?}

\vspace{0.5cm}
\large
A Stakeholder Perspective on XAI and a Conceptual Model Guiding Interdisciplinary XAI Research
            
\vspace{1.5cm}

\textbf{Markus Langer, Daniel Oster, Timo Speith\\ Holger Hermanns, Lena K{\"a}stner, Eva Schmidt, Andreas Sesing, \& Kevin Baum}

\vfill
\end{center}
\normalsize
\noindent This is a preprint version of the  paper \enquote{What do we want from Explainabile Artificial Intelligence (XAI)? -- A stakeholder perspective on XAI and a conceptual model guiding interdisciplinary XAI research.} accepted at Artificial Intelligence. The published version might differ from this version. Please cite this as: Langer M., Oster, D., Speith, T., Hermanns, H., K{\"a}stner, L., Schmidt, E., Sesing, A., \& Baum, K. (2021). What do we want from Explainabile Artificial Intelligence (XAI)? -- A stakeholder perspective on XAI and a conceptual model guiding interdisciplinary XAI research. \emph{Artificial Intelligence}. \href{https://doi.org/10.1016/j.artint.2021.103473}{doi: 10.1016/j.artint.2021.103473}

\vfill

\begin{mdframed}[outerlinecolor=black,outerlinewidth=2pt,linecolor=cccolor,middlelinewidth=3pt,roundcorner=10pt]
\begin{center}
  © 2021. This manuscript version is made available under the \href{http://creativecommons.org/licenses/by-nc-nd/4.0/}{Creative Commons Attribution-NonCommercial-NoDerivatives 4.0 International (CC-BY-NC-ND 4.0) License}.\\[2ex]
  
    \includegraphics[scale=2]{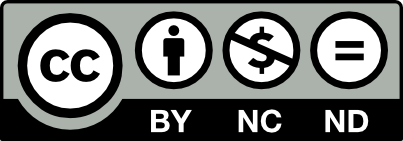}
  \end{center}
\end{mdframed}
\end{titlepage}

\begin{frontmatter}

\title{What Do We Want From Explainable Artificial Intelligence (XAI)?\\ A Stakeholder Perspective on XAI and a Conceptual Model Guiding Interdisciplinary XAI Research}\tnotetext[t1]{Work on this paper was funded by the Volkswagen Foundation grants AZ 95143, 98509, 98510, 98511, 98512, 98513, and 98514 \href{https://explainable-intelligent.systems}{\enquote{Explainable Intelligent Systems}} (EIS), by the DFG grant 389792660 as part of \href{https://perspicuous-computing.science}{TRR~248}, and by the ERC Advanced Grant 695614 (\href{https://powver.org/}{POWVER}). The authors thank the anonymous reviewers for feedback on initial drafts of this paper.}

\author{Markus Langer\fnref{first}}
\address{Department of Psychology, Saarland University, Saarbrücken, Germany}

\author{Daniel Oster\fnref{first}}
\author{Lena K{\"a}stner}
\address{Institute of Philosophy, Saarland University, Saarbrücken, Germany}

\author{Timo Speith\fnref{first}}
\author{Kevin Baum}
\address{Institute of Philosophy, Saarland University, Saarbrücken, Germany\\ Department of Computer Science, Saarland University, Saarbrücken, Germany}

\author{Holger Hermanns}
\address{Department of Computer Science, Saarland University, Saarbrücken, Germany\\ Institute of Intelligent Software, Guangzhou, China}

\author{Eva Schmidt}
\address{Institute of Philosophy and Political Sciences, Technical University Dortmund, Germany}

\author{Andreas Sesing}
\address{Institute of Legal Informatics, Saarland University, Saarbrücken, Germany}

\fntext[first]{Markus Langer, Daniel Oster, and Timo Speith have contributed equally to this article and share the first authorship.}

\begin{abstract}
Previous research in Explainable Artificial Intelligence (XAI) suggests that a main aim of explainability approaches is to satisfy specific interests, goals, expectations, needs, and demands regarding artificial systems (we call these \emph{\denquote{stakeholders' desiderata}}) in a variety of contexts. However, the literature on XAI is vast, spreads out across multiple largely disconnected disciplines, and it often remains unclear \emph{how} explainability approaches are supposed to achieve the goal of satisfying stakeholders' desiderata. This paper discusses the main classes of stakeholders calling for explainability of artificial systems and reviews their desiderata. We provide a model that explicitly spells out the main concepts and relations necessary to consider and investigate when evaluating, adjusting, choosing, and developing explainability approaches that aim to satisfy stakeholders' desiderata. This model can serve researchers from the variety of different disciplines involved in XAI as a common ground. It emphasizes where there is interdisciplinary potential in the evaluation and the development of explainability approaches. 
\end{abstract}

\end{frontmatter}


\section{Introduction}

\subsection{Background, Motivation, and Related Work}

Explainable Artificial Intelligence (XAI) is -- once again \cite{Brock2018, Clancey1983, Swartout1983, Johnson1993} -- a burgeoning multidisciplinary area of research. In general, XAI can be perceived as the topic or research field concerned with developing approaches to explain and make artificial systems understandable to human stakeholders \cite{Biran2017, Miller2019, Mittelstadt2019}. 

This puts several central aspects into the focus of XAI research. First, artificial systems are the primary objects of investigation. Such systems can range from systems following a predefined set of rules, to expert and knowledge-based systems, to systems relying on machine learning. Insights from XAI research become important when these systems are too complex to allow for human oversight or are inherently opaque, which precludes human insight \cite{Burrell2016}.
Second, this view on XAI emphasizes the importance of approaches that enable or provide insights into artificial systems, their functioning, and their outputs. These approaches (we call them \emph{\denquote{explainability approaches}}) encompass methods, procedures, and strategies to provide explanatory information helping us to better understand artificial systems. 
Third, there is a decisive need for XAI because there are human stakeholders (e.g., users, developers, regulators)\footnote{Generally, we speak of single stakeholders here. Since we cannot consider each stakeholder individually, we treat them as representative members of specified stakeholder classes.} 
whose interests, goals, expectations, needs, and demands regarding artificial systems (e.g., to have fair or trustworthy systems \cite{DoshiVelez2017, EUHLEGOAI}) call for greater understandability of artificial systems. We call such conglomerations of stakeholders' interests, goals, expectations, needs, and demands regarding artificial systems \emph{\denquote{stakeholders' desiderata}}.

A large part of previous XAI research was mainly concerned with developing new explainability approaches without evaluating whether these methods are useful to satisfy stakeholders' desiderata (except maybe the desiderata of developers) \cite{DoshiVelez2017, Lipton2018, Adadi2018, Nunes2017}. In fact, only a minority of papers concerned with explainability approaches also evaluated the proposed methods \cite{Adadi2018, Nunes2017}. In contrast, nowadays an increasing number of researchers strongly suggest putting human stakeholders in the center of attention when evaluating and developing explainability approaches. For instance, researchers have proposed to comprehensively examine the perspectives of all stakeholders involved in the discussions around XAI (e.g., \cite{Arrieta2020, Felzmann2019, Gilpin2018b}) or they have introduced evaluation methods and metrics to systematically and empirically investigate the effects of explainability approaches on human stakeholders and their desiderata (e.g., \cite{DoshiVelez2017, Hoffman2018}).

This paper reinforces and extends the focus on human stakeholders as well as on the development and evaluation of explainability approaches, and provides three main contributions. First, we propose that when evaluating, adjusting, choosing, and developing explainability approaches, research needs to pay more attention to stakeholders' specific desiderata in given contexts. This is crucial as the success of explainability approaches depends on how well they satisfy these desiderata. Current measures and metrics focus on how well explainability approaches calibrate trust or how much they increase human-machine performance (see, e.g., \cite{Hoffman2018}). However, these are just two of many desiderata driving XAI research and although there is research that investigates which classes of stakeholders hold essential desiderata for XAI (e.g., \cite{Preece2018, Arrieta2020, Felzmann2019, Weller2019}), there is a lack of research identifying, defining, and empirically investigating these desiderata, let alone research that links them to explainability approaches suitable for their satisfaction. Our paper identifies desiderata of different classes of stakeholders and calls for systematic empirical research investigating how explainability approaches, through the facilitation of understanding, lead to the satisfaction of these desiderata. 

Second, we emphasize the central role of understanding as a path through which explainability approaches satisfy stakeholders' desiderata. Although research has highlighted the importance of human understanding to XAI (e.g., \cite{Hoffman2018, Paez2019}), understanding sometimes seems to be considered as just one of many important outcomes of explainability approaches \cite{Cheng2019}. We claim that increasing human understanding is not just one of many important effects of explainability approaches, but crucial for the satisfaction of desiderata in general. For this reason, we introduce a model that emphasizes the critical importance of human understanding as a mediator between explainability approaches and the satisfaction of desiderata (for a related model focused on user performance, see Hoffman et al. \cite{Hoffman2018}). 

Third, we propose that our model can be used to guide evaluating, adjusting, choosing, and developing explainability approaches. In particular, our model highlights the main concepts and their relations of how explainability approaches are supposed to lead to the satisfaction of stakeholders' desiderata. Clearly defining, analyzing, and capturing these concepts, as well as clarifying their relations are central for the systematic evaluation of the success of explainability approaches, as this helps to identify potential reasons for why an explainability approach did not satisfy given desiderata. Similarly, considering these concepts and their relations is crucial for the choice between different explainability approaches or for the successful development of such approaches, because this supports the derivation of requirements for an explainability approach that has the potential to satisfy stakeholders' desiderata. Furthermore, our model is useful to detect where input from disciplines outside of computer science (e.g., psychology, philosophy, law, sociology; \cite{Abdul2018}) is crucial when evaluating or developing explainability approaches. Thus, our model serves to identify interdisciplinary potential and is aimed to establish a common ground for different disciplines involved in XAI. Overall, the current paper is intended for an interdisciplinary readership interested in XAI. 

\subsection{A Conceptual Model of the Relation Between Explainability Approaches and Stakeholders' Desiderata}

\begin{figure}[ht]
\centering
\begin{tikzpicture}[scale=0.75,transform shape, align=center,
    block/.style={
      rectangle,
      draw=black,
      thick,
      text width=7em,
      align=center,
      rounded corners,
      minimum height=3.5em
    },
    block1/.style={
      rectangle,
      draw=black,
      thick,
      text width=12em,
      align=center,
      rounded corners,
      minimum height=1.5em
    },
     block2/.style={
      rectangle,
      draw=black,
      thick,
      text width=6em,
      align=center,
      rounded corners,
      minimum height=1.5em
    },
      block3/.style={
      rectangle,
      draw=black,
      thick,
      text width=10em,
      align=center,
      rounded corners,
      minimum height=1.5em
    },
      block4/.style={
      rectangle, 
      rounded corners,
      draw=black,
      thick,
      text width=18em,
      align=center,
      minimum height=6.5em
    },
    line/.style={
      draw,thick,
      -latex',
      shorten >=2pt
    },
    cloud/.style={
      draw=red,
      thick,
      ellipse,
      fill=red!10,
      text width=12em,
      minimum height=1em
    }
  ]

 \node[block1]   at (0,7)   (des1) {Desiderata Satisfaction};
 
 \node[block4, below of = des1, node distance = 0.725cm]        (block) {};
 
 \node[below of = des1, node distance = 1.5cm]        (h2) {};
 
 \node[block2, left of = h2, node distance = 2cm]        (ep1) {epistemic};
 \node[block2, right of = h2, node distance = 2cm]        (sub1) {substantial};
 
\node[block, left of = block, node distance = 6cm]  (und1) {Stakeholders' Understanding};
 
 \node[block, left of = und1, node distance = 4.5cm] (exp1) {Explanatory Information};

 \node[block, left of = exp1, node distance = 4.5cm]     (xai1) {Explainability Approach};
 
 \node[block3, above of = block, node distance = 2.5cm] (sta) {Status of \mbox{Desiderata} Satisfaction};

 \draw[->] (und1) -- (block) node[midway, below] (aff1) {affects};
 \draw[->] (exp1) -- (und1) node[midway, below] (fac1) {facilitates};
 \draw[->] (xai1.east) -- (exp1.west) node[midway, below] (fac2) {provides};
 
 \node[block2, above of = fac1, node distance = 2cm] (con1) {Context};
 

 \draw[->] ($(con1.south)+(0.5,0)$) to[out=315, in=90] (aff1);
 \draw[->] (con1) -- (fac1);
 
 \node[fill=white] at ($(con1.south)+(0.5,-0.35)$) (mod) {moderates};
 
 \draw[-] (des1) -- (ep1);
 \draw[-] (des1) -- (sub1);
 \draw[->] (ep1) -- (sub1) node[midway, above] (ena) {enables};

\draw[decorate,decoration={brace,amplitude=12pt}] ($(xai1.north west)+(0,3)$) -- ($(und1.north east)+(0,3)$) node[midway, above, yshift=12pt] (xxx2) {Explanation Process};

\draw[decorate,decoration={brace,amplitude=12pt}] ($(block.north west)+(0,2.5)$) -- ($(block.north east)+(0,2.5)$) node[midway, above, yshift=12pt] (xxx3) {Desiderata};

\draw[->] (xxx3) -- (xxx2) node[midway, above] (mot) {motivate and guide};

\draw[-] (sta) to (block);
 \draw[->] (sta.west) to[out=170, in=330] node[midway, above] (feb2) {feeds back} ($(xxx2.south)+(0,-0.3)$); 
 
\end{tikzpicture}
\vspace{-3ex}
\caption{Our proposed model of how explainability approaches relate to the satisfaction of stakeholders' desiderata.}
\label{fig:ourxai}
\vspace{-0.5em}
\end{figure}
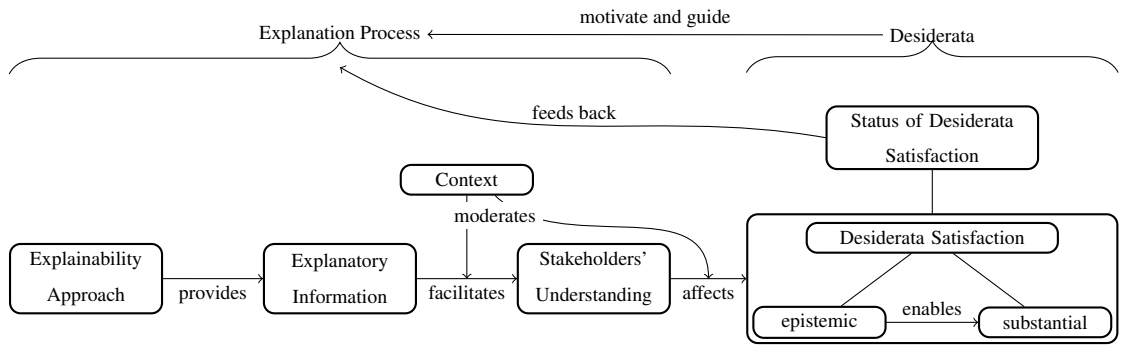

For the purposes of this paper, we introduce a conceptual model (see Figure \ref{fig:ourxai}) 
that organizes and makes explicit the central concepts of how explainability approaches relate to the satisfaction of stakeholders' desiderata, as well as the relations between these concepts. The main concepts in this model are: \denquote{explainability approach}, \denquote{explanatory information}, \denquote{stakeholders' understanding}, \denquote{desiderata satisfaction}, and \denquote{(given) context}.

The overall idea of our model is that the success of an explainability approach depends on the satisfaction of stakeholders' desiderata (consisting of the \emph{substantial} and the \emph{epistemic} facet of desiderata satisfaction, see Section \ref{sec:facets}). Desiderata satisfaction, thus, motivates an explanation process including explainability approaches, explanatory information, and stakeholders' understanding. Specifically, in the explanation process we assume that explainability approaches provide explanatory information to human stakeholders. Human stakeholders engage with the information to facilitate their understanding of an artificial system, its functioning, and outputs. As a consequence, the adjusted understanding of the stakeholders affects the extent to which their desiderata are satisfied. The context in which the human stakeholder and the artificial system operate and interact affects the relations between the other concepts (i.e., influences the relation between explanatory information and stakeholder understanding as well as the relation between understanding and desiderata satisfaction). Identifying, defining, as well as capturing and empirically examining the concepts and their relations should guide evaluating, adjusting, choosing, and developing of explainability approaches that aim to satisfy stakeholders' desiderata.

With a focus on stakeholders' desiderata, the following sections will elaborate on the model's concepts and their relations in more detail, as well as explicate shortcomings of the current view on these concepts and their relations. This paper is structured as follows. We will start on the right side of Figure \ref{fig:ourxai} and will continue to work backwards from stakeholders' desiderata. In Section \ref{sec:des}, we will describe different classes of stakeholders and provide examples of their pertinent desiderata. We will elaborate on the central role of understanding for desiderata satisfaction in Section \ref{sec:desrun}. Evoking understanding, in turn, requires explanatory information, as we will illuminate in Section \ref{sec:unrei}. Section \ref{sec:eirea} will shed light on the connection between explainability approaches and explanatory information. Throughout these sections, we will point towards interdisciplinary potential that becomes apparent with the transition from one concept to the next in our model. In Section \ref{sec:use}, we will, then, exemplify how our model can be used to evaluate, adjust, choose, and develop explainability approaches.

\section{Stakeholders' Desiderata}
\label{sec:des}

Stakeholders' desiderata are one, if not \emph{the}, reason for the rising popularity of XAI (see also \cite{Arrieta2020, Weller2019}). Since stakeholders in combination with their concrete desiderata motivate, guide, and affect the explanation process depicted in Figure \ref{fig:ourxai}, we propose that identifying and clarifying desiderata of the various classes of stakeholders related to artificial systems is a crucial first step when evaluating, adjusting, choosing, and developing explainability approaches for an artificial system in a given context.

\subsection{Stakeholder Classes}

The need for explainability starts with the increasing societal impact of artificial systems and the fact that many such systems still have to be operated by humans or affect human lives. This indicates that there are various groups of people with different interests in the explainability of artificial systems: people operate systems, try to improve them, are affected by decisions based on their output, deploy systems for everyday tasks, and set the regulatory frame for their use. These people are commonly called \emph{stakeholders}.\footnote{According to the Merriam-Webstar dictionary, a stakeholder is, among others, someone \enquote{who is involved in or affected by a course of action} \cite{MerriamStakeholder}. For this reason, we use this as a general term, but refer to specific stakeholder classes where appropriate.}

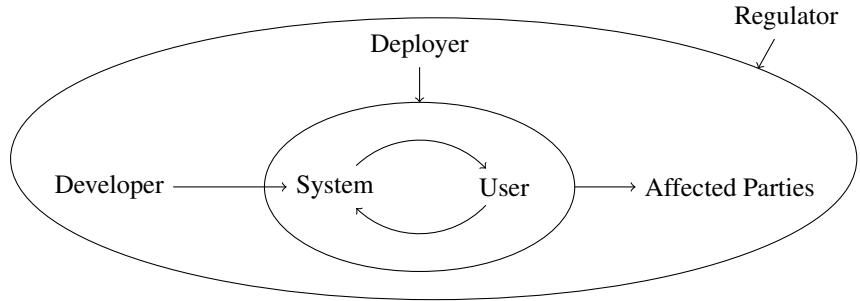
\begin{figure}[ht]
\centering
\begin{tikzpicture}[scale=0.75]
    
    \node[] at (-4, 0)  (dev) {Developer};
    \node[] at (0, 0)   (sys) {System};
    \node[] at (1.5, 2.5)(dep) {Deployer};
    \node[] at (3, 0)   (usr) {User};
    \node[] at (7, 0)   (aff) {Affected Parties};
    
    \draw[->] (dev) -- (sys)  node[midway, left] (let) {};
    \draw[->, bend left = 45] (sys) to node[midway, left] (x1) {} (usr);
    \draw[->, bend left = 45] (usr) to node[midway, left] (x2) {} (sys);
    \draw[->] (4.25,0) to node[midway, left] (x3) {} (aff);
    
    \draw (1.5,0) ellipse (2.75cm and 1.5cm);
    
    \draw[->] (dep) -- (1.5, 1.5);
    
    \draw (1.75, 0.5) ellipse (7.5cm and 2.5cm);
    
    \node[] at (8, 3) (reg) {Regulator};
    
    \draw[->] (reg) -- (7.5, 2.1);
    
\end{tikzpicture}
\vspace{-1ex}
\caption{The different classes of stakeholders associated with artificial systems and their relations.}
\label{fig:stakeholder}
\vspace{-0.5em}
\end{figure}

Previous research has discussed varying classes of stakeholders in the context of XAI. For instance, Preece et al. \cite{Preece2018} distinguish between four main classes of stakeholders: \emph{developers}, \emph{theorists}, \emph{ethicists}, and \emph{users}. Arrieta et al.~\cite{Arrieta2020} categorize the main classes of stakeholders into domain experts/users, data scientists/developers/product owners, users affected by model decisions, managers/executive board members, and regulatory entities (see also \cite{Weller2019, Hind2019, Felzmann2019}). We follow Arrieta et al.~and distinguish five classes of stakeholders: \emph{users}, \emph{(system) developers}, \emph{affected parties}, \emph{deployers}, and \emph{regulators} (see Figure \ref{fig:stakeholder}).

Clearly, one person can be a member of several stakeholder classes. A user, for instance, can be affected by the outputs of the system she operates. Additionally, these are just prototypical classes of stakeholders and more fine-grained distinctions into sub-classes of stakeholders are possible \cite{Felzmann2019}. For example, there is not one prototypical developer, but developers differ in their expertise and in other factors (e.g., personality). A novice developer may have different desiderata than an expert. In a similar way, a lay user's desiderata might differ from those of an expert user (more on this in Section \ref{sec:conexu}). Moreover, we want to emphasize that this list of stakeholders is not necessarily exhaustive because our distinction is based on previous research that mainly comes from a computer scientific background. Thus, it might neglect other classes of stakeholders.

\subsection{Exemplary Stakeholders' Desiderata}

The desiderata arising from the five classes of stakeholders are diverse. Based on a term search, we identified more than 100 peer-reviewed journal and conference publications that postulate XAI as being indispensable when it comes to satisfying the different desiderata (see Table \ref{tab:desiderata}). 

\newgeometry{left=1.5cm, bottom=2.5cm}
\begin{landscape}
\begin{table}[!ht]
    \footnotesize
    \centering
    \scalebox{0.825}{
    \begin{tabular}{l | p{9cm} | l | p{8cm} | l | l | p{2.2cm}}
        \hline
        Desideratum & Tentative Description & Stakeholder & Without Empirical (Emp.) Investigation & Emp. Evidence & Emp. Mixed & Emp. Inconclusive\\
        \hline
        
        Acceptance & Improve acceptance of systems & Deployer, Regulator & \cite{Anjomshoae2019b, Atzmueller2019, Baaj2019, Balog2019, Binns2018, Biran2017, Chakraborti2019, Chen2019, Cotter2017,  Darlington2013, Ehrlich2011, Felzmann2019, Freitas2014, Gregor1999, Hois2019, Kizilcec2016, Nagulendra2016, Nunes2017, Papenmeier2019, Pierrard2019, Putnam2019, Rader2018, Rosenfeld2019, Sato2019, Vig2009, Watts2019, Zhou2019b}
        & \cite{Herlocker2000} & \cite{Cramer2008} & \\
        
        Accountability & Provide appropriate means to determine who is accountable & Regulator & \cite{Abdul2018, Binns2018, Byrne2019, DeLaat2018, Felzmann2019, Floridi2018, Gilpin2018b, Lepri2018, Mathews2019, Mittelstadt2016, Mittelstadt2019, Paez2019, Pieters2011, Ras2018, Riedl2019, Robbins2019, Sheh2017, Sheh2018, Sokol2020, Sokol2020b, Sridharan2019, Vellido2019, Wang2019}
        & \cite{Lee2019, Rader2018} & & \\
        
        Accuracy & Assess and increase a system's predictive accuracy & Developer & \cite{Anjomshoae2019b, Biran2017, Byrne2019, Darlington2013, Doran2018, Kizilcec2016, Krishnan2019, Mathews2019, Peddoju2019, Rajani2017, Vig2009, Zhou2019}
        & & & \\
        
        Autonomy & Enable humans to retain their autonomy when interacting with a system & User & \cite{Anjomshoae2019, Felzmann2019, Floridi2018, Fox2017, Jasanoff2017, Mittelstadt2016, Pieters2011, Robbins2019, Rosenfeld2019}
        & & & \\
        
        Confidence & Make humans confident when using a system & User & \cite{Anjomshoae2019, Arrieta2020, Balog2019, Biran2017, Cotter2017, Doran2018, Ehrlich2011, Felzmann2019, Freitas2014, Friedrich2011, Gilpin2018b, Gregor1999, Herlocker2000, Holzinger2019, Kizilcec2016, Nagulendra2016, Pieters2011, Ras2018, Sevastjanova2018, Sormo2005, Zerilli2019, Zhou2019, Zhou2019b}
        & & & \cite{Lucic2019} \\
        
        Controllability & Retain (complete) human control concerning a system & User & \cite{Abdul2018, Adadi2018, Anjomshoae2019, Herlocker2000, Miller2019, Mittelstadt2016, Nunes2017, Robbins2019, Rosenfeld2019}
        & \cite{Nagulendra2016} & \cite{Rader2018} & \\
        
        Debugability & Identify and fix errors and bugs & Developer & \cite{Adadi2018, Anjomshoae2019, Anjomshoae2019b, Biran2017, Byrne2019, Chakraborti2019, Dam2018, Darlington2013, DeWinter2010, Gilpin2018b, Juozapaitis2019, Michael2019, Mittelstadt2016, Mittelstadt2019, Nunes2017, Ras2018, Rosenfeld2019, Sheh2017, Sheh2018, Sokol2018, Sokol2020, Sokol2020b, Wang2019, Watts2019, Wicaksono2017, Zhou2019b}
        & \cite{Eiter2019, Kulesza2011, Lucic2019} & & \\
        
        Education & Learn how to use a system and system's peculiarities & User & \cite{Anjomshoae2019, Chen2019, Darlington2013, Gregor1999, Herlocker2000, Hind2019, Hoffman2018b, Miller2019, Nagulendra2016, Nothdurft2013, Nunes2017, Sormo2005, Wicaksono2017}
        & & &\\
        
        Effectiveness & Assess and increase a system's effectiveness; work effectively with a system & Developer, User & \cite{Anjomshoae2019, Atzmueller2019, Brinton2017, Ehrlich2011, Friedrich2011, Gregor1999, Herlocker2000, Holzinger2019, Nagulendra2016, Nunes2017, Putnam2019, Sridharan2019, Tintarev2007, Wang2019, Weber2019, Zhou2019b}
        & \cite{Balog2019, Chen2019, Vig2009} & & \\
        
        Efficiency & Assess and increase a system's efficiency; work efficiently with a system & Developer, User & \cite{Anjomshoae2019, Balog2019, Brinton2017, Holzinger2019, Mathews2019, Nagulendra2016, Nunes2017, Tintarev2007, Weber2019}
        & \cite{Chen2019} & & \\
        
        Fairness & Assess and increase a system's (actual) fairness & Affected, Regulator &  \cite{Abdul2018, Adadi2018, Arrieta2020, Burrell2016, Byrne2019, Cotter2017, Doran2018, Floridi2018, Gilpin2018, Gilpin2018b, Hind2019, Ho2020, Hohman2019, Holzinger2019, Kizilcec2016, Krishnan2019, Mathews2019, Michael2019, Mittelstadt2016, Nunes2017, Papenmeier2019, Ras2018, Robbins2019, Rosenfeld2019, Sokol2018, Sokol2020, Sokol2020b, Veale2017, Wang2019, Zednik2019, Zerilli2019}
        & & \cite{Binns2018, Lee2019, Rader2018} & \\
        
        Informed Consent & Enable humans to give their informed consent concerning a system's decisions & Affected, Regulator & \cite{Felzmann2019, Mittelstadt2016, Pieters2011}
        & & & \\
        
        Legal Compliance & Assess and increase the legal compliance of a system & Deployer &  \cite{Abdul2018, Adadi2018, Anjomshoae2019, Atzmueller2019, Baaj2019, Binns2018, Biran2017, Chakraborti2019, Cheng2019, Doran2018, Ehrlich2011, Felzmann2019, Fox2017, Freitas2014, Gilpin2018b, Goodman2017, Hind2019, Hoffman2018b, Hohman2019, Jasanoff2017, Krishnan2019, Lee2019, Mathews2019, Mittelstadt2016, Mittelstadt2019, Ras2018, Rosenfeld2019, Sheh2018, Sklar2018, Sokol2018, Sokol2020, Vellido2019, Watts2019, Zednik2019, Zerilli2019}
        & & & \\
        
        Morality/Ethics & Assess and increase a system's compliance with moral and ethical standards & Affected, Regulator & \cite{Adadi2018, Arrieta2020, Binns2018, Doran2018, Ehrlich2011, Floridi2018, Gilpin2018b, Hois2019, Holzinger2019, Mathews2019, Miller2019, Pieters2011, Ras2018, Robbins2019, Rosenfeld2019, Watts2019}
        & & & \\
        
        Performance & Assess and increase the performance of a system & Developer & \cite{Anjomshoae2019, Atzmueller2019, Byrne2019, Cotter2017, Cramer2008, Darlington2013, Felzmann2019, Gilpin2018, Gregor1999, Hoffman2018b, Kizilcec2016, Papenmeier2019, Rader2018, Rajani2017, Sklar2018, Vellido2019, Zednik2019, Zhou2019b}
        & & \cite{Ehrlich2011, Lage2019} & \cite{Herlocker2000}\\
        
        
        Privacy & Assess and increase a system's privacy practices & User & \cite{Arrieta2020, Gilpin2018b, Hohman2019, Holzinger2019}
        & & & \\
        
        
        Responsibility & Provide appropriate means to let humans remain responsible or to increase perceived responsibility & Regulator & \cite{Dahl2018, Miller2019, Mittelstadt2016, Nunes2017, Paez2019, Pieters2011, Rader2018, Robbins2019}
        & \cite{Lee2019} & & \\
        
        Robustness & Assess and increase a system's robustness (e.g., against adversarial manipulation) & Developer &  \cite{Arrieta2020, Biran2017, Ghosh2019, Mathews2019}
        & & & \\
        
        Safety & Assess and increase a system's safety & Deployer, User & \cite{Anjomshoae2019, Doran2018, Ghosh2019, Holzinger2019, Krishnan2019, Ras2018, Rosenfeld2019}
        & & & \\
        
        Satisfaction & Have satisfying systems & User & \cite{Balog2019, Biran2017, Darlington2013, Felzmann2019, Gregor1999, Hind2019, Holzinger2019, Nagulendra2016, Nunes2017, Rader2018, Sevastjanova2018, Sklar2018, Tintarev2007, Vig2009}
        & & & \cite{Chen2019} \\
        
        Science & Gain scientific insights from the system & User & \cite{Adadi2018, Arrieta2020, Biran2017, Freitas2014, Ho2020, Hois2019, Krishnan2019, Mittelstadt2019, Robbins2019, Rosenfeld2019, Sokol2018, Sokol2020, Stuart2019, Vellido2019, Wang2019, Watts2019, Zednik2019}
        & & & \\
        
        Security & Assess and increase a system's security & All & \cite{Holzinger2019, Mathews2019, Pieters2011, Zednik2019}
        & & & \\
        
        Transferability & Make a system's learned model transferable to other contexts & Developer & \cite{Arrieta2020, DeWinter2010, Hois2019}
        & & & \\
        
        Transparency & Have transparent systems & Regulator & \cite{Anjomshoae2019, Anjomshoae2019b, Atzmueller2019, Binns2018, Clos2017, Darlington2013, Ehrlich2011, Fox2017, Gilpin2018b, Herlocker2000, Hois2019, Holzinger2019, Mathews2019, Nagulendra2016, Nothdurft2013, Nunes2017, Pieters2011, Rosenfeld2019, Sheh2018, Sokol2020, Sormo2005, Tintarev2007, Vig2009, Wang2019, Zhu2018}
        & \cite{Balog2019, Chen2019, Cramer2008, Rader2018} & & \\
        
        Trust & Calibrate appropriate trust in the system & User, Deployer & \cite{Abdul2018, Adadi2018, Anjomshoae2019, Anjomshoae2019b, Atzmueller2019, Baaj2019, Balog2019, Binns2018, Biran2017, Brinton2017, Byrne2019, Clinciu2019, Clos2017, Cotter2017, Dam2018, Darlington2013, Doran2018, Ehrlich2011, Felzmann2019, Floridi2018, Fox2017, Freitas2014, Friedrich2011, Gilpin2018, Gilpin2018b, Gregor1999, Henin2019, Herlocker2000, Hoffman2018b, Hohman2019, Hois2019, Holzinger2019, Krishnan2019, Lage2019, Madumal2019b, Mathews2019, Michael2019, Miller2019, Mittelstadt2016, Mittelstadt2019, Nunes2017, Olson2019, Paez2019, Peddoju2019, Pieters2011, Putnam2019, Rajani2017, Ras2018, Riedl2019, Rosenfeld2019, Sato2019, Sevastjanova2018, Sheh2018, Sokol2018, Sokol2020, Stuart2019, Tintarev2007, Vig2009, Wang2019, Watts2019, Weber2019, Wicaksono2017, Zednik2019, Zeng2018, Zerilli2019, Zhou2019, Zhu2018}
        & \cite{Chakraborti2019, Nagulendra2016}
        & \cite{Kizilcec2016, Papenmeier2019, Zhou2019b}
        & \cite{Chen2019, Cheng2019, Cramer2008, Lucic2019, Madumal2019a, Nothdurft2013}\\
        
        Trustworthiness & Assess and increase the system's trustworthiness & Regulator & \cite{Adadi2018, Anjomshoae2019, Dahl2018, Darlington2013, Doran2018, Friedrich2011, Ghosh2019, Hoffman2018b, Kizilcec2016, Mittelstadt2019, Pierrard2019, Robbins2019, Tintarev2007}
        & & & \\
        
        Usability & Have usable systems & User & \cite{Abdul2018, Balog2019, Dahl2018, Darlington2013, Gregor1999, Henin2019, Holzinger2019, Nunes2017, Pieters2011, Rader2018, Sheh2018, Sokol2020, Zhu2018}
        & & & \\
        
        Usefulness & Have useful systems & User & \cite{Ehrlich2011, Friedrich2011, Gregor1999, Nunes2017, Rader2018, Sokol2020}
        & \cite{Chen2019, Sato2019} & & \\
        
        Verification & Be able to evaluate whether the system does what it is supposed to do & Developer & \cite{Arrieta2020, Cotter2017, Darlington2013, Gilpin2018b, Mathews2019, Mittelstadt2016, Mittelstadt2019, Paez2019, Rader2018, Rosenfeld2019, Sheh2018, Sormo2005, Vellido2019}
        & & & \\
        
        \hline
\end{tabular}}
\centering
    \caption{An exemplary list of desiderata, stakeholders holding these desiderata, and sources that claim, propose, or show that XAI-related research (e.g., on explainability approaches) and its findings and outputs can contribute to the satisfaction of these desiderata. We classify the sources into those that provide no empirical investigation of their claims, those that show empirical evidence (e.g., an explainability approach affected a desideratum's satisfaction), those that provide mixed empirical evidence (e.g., for some explainability approaches there are positive effects on a given desideratum's satisfaction, whereas for others there are no effects), and those that present inconclusive empirical evidence (e.g., the effect of an explainability approach on a desideratum's satisfaction was not significant).}
    \label{tab:desiderata}
\end{table}
\end{landscape}
\restoregeometry

Table \ref{tab:desiderata} presents exemplary desiderata that we have extracted from this literature review. Each row contains a desideratum, partnered with sources that claim, propose, or show that XAI-related research (e.g., on explainability approaches) and its findings and outputs can contribute to the satisfaction of this desideratum. Furthermore, this table presents stakeholder classes that may be most prone to have one or more of these desiderata. Whenever we could not extract stakeholder classes from the respective papers, we did our own (mostly common-sense) mappings. Notably, most of the sources we present in this table only claim that XAI-related research can contribute to satisfy the respective desiderata with only a subset of these papers (for instance, \cite{Balog2019, Chen2019, Sato2019}) providing empirical evidence for their claims (e.g., regarding the mapping of desiderata to stakeholder classes or regarding the relation of explanatory information and desiderata).
In what follows, we present two important exemplary desiderata for each class of stakeholders.

\emph{Users.}
Most papers concerning stakeholders in XAI have this class of stakeholders in common (see, e.g., \cite{Arrieta2020, Preece2018, Weller2019}). Among others, users take into account recommendations of artificial systems to make decisions \cite{Hind2019}. Some prototypical members of this stakeholder class are medical doctors, loan officers, judges, or hiring managers. Usually, users are no experts regarding the technical details and the functioning of the systems they use. However, they can work more effectively if they form adequate expectations concerning the systems' functioning. In case they cannot do so, and in cases where their expectations are violated, they need information that goes beyond the knowledge of purely operating the system. This motivates at least the following two central desiderata of users: \emph{usability} \cite{Abdul2018, Gregor1999} and \emph{trust} \cite{Ribeiro2016, Gilpin2018, Binns2018}.

In many cases, a system is more usable if it offers meaningful information alongside its outputs. This information can help users to adequately link their knowledge and assessment of a given situation to the information used by a system, can help them to make decisions more quickly, or to increase decision quality \cite{Endsley2017}. All of this can contribute to the usefulness (another important desideratum of users) of a system and is important in high-stakes scenarios where a user decides on the basis of a system's recommendations.

This is closely linked to the desideratum of adequately calibrating trust in systems. Both undertrust and overtrust can negatively affect the appropriate use of systems \cite{Lee2004}. In the case of undertrust, users may constantly try to supervise a system's behavior or even attempt to intervene in a system's processes, thereby undermining the effectiveness of the human-system interaction \cite{Parasuraman1997}. In the case of overtrust, people may use a system without questioning its behavior \cite{Hoff2014, Parasuraman1997, Parasuraman2010}. This can again decrease the effectiveness of the human-system interaction, as humans rely on the system's outputs even in situations where they should challenge them \cite{Kunze2019, Lee2004}. Explainability approaches have the potential to provide means to let users adequately calibrate their trust in artificial systems \cite{Hoffman2018}.

\emph{Developers.} 
Individuals who design, program, and build artificial systems are the developers. Naturally, they count as a class of stakeholders, as without them the systems would not exist in the first place. Generally, developers have a high expertise concerning the systems and an interest in improving them.

An especially important desideratum of developers is \emph{verification}, that is, to check whether a system works as intended \cite{Samek2017, Montavon2018, Becker2018, Mittelstadt2019}. There are many ML-based classifiers that consider, for instance, irrelevant inputs as relevant (see, e.g., \cite{Lapuschkin2016, Caruana2015}). Increasing insights into the system's decision-making processes by using certain explainability approaches can help developers to recognize and correct such mistakes. Accordingly, there are cases where XAI contributes to determine whether a system works as intended and, thus, explainability approaches can support verification of the system.

Another important desideratum for developers is \emph{performance}. There are many ways in which a system can achieve a better performance. For example, the predictive accuracy of an ML algorithm can be seen as a performance measure. Although there are some claims that explainability and accuracy are difficult to combine \cite{Papenmeier2019, Rosenfeld2019}, there is also the opposite view, which sees XAI as a way to actually make systems more accurate and, in particular, to help developers estimate system accuracy \cite{Krishnan2019, Zhou2019}. By means of getting information of what led to a system's outcomes, developers can detect underrepresented or erroneous training data and, thus, fine-tune the learning process to achieve higher accuracy. Another way in which performance can be understood is user-system interaction. The better users can interact with a system, the better they, the system, and the combination of user and system perform. To this end, insights about a system, its functioning, and its outputs are a fruitful way to improve user-system performance \cite{Hoffman2018, Zhou2019b, Arrieta2020}.

\emph{Affected Parties.}
The influence of artificial systems is constantly growing and decisions about people are increasingly automated -- often without their knowing. Affected parties are such (groups of) people in the scope of a system's impact. They are stakeholders, as for them much hinges on the decision of an artificial system. Patients, job or loan applicants, or defendants at court are typical examples of this class.

Crucial desiderata of affected parties are \emph{fairness} \cite{Lipton2018, Samek2017, DoshiVelez2017, Mittelstadt2019} and \emph{morality/ethics} \cite{Lipton2018, Miller2019, Baum2018}. These desiderata are closely related. If a system is fair, for instance, the influence of protected attributes (e.g., gender or ethnicity) is adequately limited or controlled in the systems' decision-making processes. In the case of ethical systems, their decision-making processes rely on morally permissible considerations (e.g., according to certain moral theories, an autonomous car in a dilemma situation should never let affected parties' age contribute to its decision-making process, see \cite{Luetge2017}).

Considerations of fairness and ethics have evolved because there is an increasing number of affected parties. This can lead to discrimination of individuals (e.g., concerning the distribution of jobs, loans, or healthcare), not on the basis of their own actions or characteristics but on the basis of actions or characteristics of social groups to which they belong (e.g., women, ethnic minorities, older people) \cite{Lepri2018}. One hope of establishing automated decision-processes was to make decisions less prone to human bias \cite{Purkiss2006}. However, it is commonly acknowledged that artificial systems can reproduce and, in this process, even intensify human biases (see, e.g., \cite{Lepri2018} and \cite{Caliskan2017}). To counteract biases, it is, therefore, crucial to enable their detection. Explainability approaches may aid in this regard by providing means to track down factors that may have contributed to unfair and unethical decision-making processes and either to eliminate such factors, to mitigate them, or at least to be aware of them.

\emph{Deployers.}
People who decide where to employ certain systems (e.g, a hospital manger decides to implement a diagnosis system in her hospital) are deployers. We count them as another class of stakeholders because their decisions influence many other classes of stakeholders. For example, users have to work with the deployed systems and, consequently, new people fall inside of the range of affected parties.

Deployers want the systems they bring into use to be \emph{accepted} \cite{Lipton2018, Guidotti2019, Gilpin2018, Biran2017}. In the eyes of deployers, the worst case in terms of acceptance is that users reject appropriately working systems so that the systems will end up never being used \cite{Venkatesh2003}. Therefore, low acceptance undermines what deployers intend to achieve when providing systems to users. Previous research claims that explainability approaches can aid in this case by providing people with more insights into systems, which can improve their acceptance \cite{Lipton2018, Guidotti2019, Ribeiro2016}.

Another desideratum of deployers is the system's \emph{legal compliance}. As deployers bear a certain degree of responsibility for systems they bring into use, they have to ensure that these systems comply with legislation. Non-discrimination and safety of a system are two important factors for its legal compliance. Explainability approaches promise to enable deployers and other stakeholders to check whether the system is indeed safe and non-discriminatory. Moreover, the European General Data Protection Regulation (GDPR) and the often discussed \emph{Right to Explanation} \cite{Goodman2017} (arguably) explicitly require explanations.

\emph{Regulators.}
Finally, there are regulators stipulating legal and ethical norms for the general use, deployment, and development of systems. This class of stakeholders occupies a somewhat extraordinary role, since they have a \denquote{watchdog} function not only with regard to systems, but to the whole interaction process of systems and the other stakeholder classes. This class consists of ethicists, lawyers, and politicians, who must have the know-how to assess, control, and regulate the whole process of using artificial systems.

Regulators call, for instance, for \emph{trustworthy} systems \cite{Lipton2018, Ribeiro2016, Gilpin2018, EUHLEGOAI, Binns2018, Guidotti2019, Miller2019, Biran2017}. However, the concept of trustworthiness is still only vaguely defined \cite{McLeod2015}. For example, the High Level Expert group on Artificial Intelligence (HLEGAI) initiated by the European Commission does not provide a common definition for trustworthiness, but it only proposes that trustworthy systems have three properties: they are lawful, ethical, and robust \cite{EUHLEGOAI}. Without examining trustworthiness more closely, the HLEGAI emphasizes the significance of trustworthy artificial systems by stating that the trustworthiness of systems is imperative for the realization of potentially vast social and economic benefits. Regulators such as the EU, as well as previous research on artificial intelligence that calls for trustworthy systems (e.g., as described in \cite{Floridi2018}), agree that explainability approaches are one central way to facilitate the trustworthiness of systems \cite{EUHLEGOAI, Floridi2018}.

\emph{Accountability} is another important desideratum of regulators \cite{Paez2019, Mittelstadt2016}. Accountability is about being able to identify who is blamable or culpable for a mistake. With increasing use of artificial systems, accountability gaps might emerge \cite{Raji2020, Matthias2004}. For instance, when the use of an artificial system harms a person, it may not be clear who is accountable, as there are many parties that may have contributed to the harm. Opaque artificial systems only amplify this issue. For example, a person acting on the outputs of a system may not (be able to) know that this output was erroneous, so blaming her for ensuing problems might inadequately ignore the contribution of the artificial systems. Overall, regulators want to avoid situations in which existing legislation is hard to apply or where no one is (or feels) accountable for a mistake. In such cases, explainability approaches may restore accountability by making errors and causes for unfavorable outcomes detectable and attributable to the involved parties.

\subsection{Interdisciplinary Potential}

Artificial systems will continue to influence humans in every part of their lives, thus it is likely that new desiderata will emerge. Further desiderata might evolve from societal, legal, political, philosophical, or psychological needs regarding artificial systems (e.g., for competence, relatedness, or autonomy; \cite{Deci2017}). For example, with artificial systems in healthcare \cite{Longoni2019} there is a pressing need for formulating relevant ethical and legal desiderata. In addition, it is also possible that explanatory information provided by artificial systems does not only aim to improve task achievement but also to entertain users \cite{Keil2006}.

We conducted a literature review to derive an overview of stakeholders' desiderata, but it will clearly be possible to extend our list in Table \ref{tab:desiderata}. In fact, further developing and refining this list of desiderata is an important point that reveals interdisciplinary potential. First, most of the sources referred to in Table \ref{tab:desiderata} only claim that these desiderata are relevant for stakeholders. There needs to be a more thorough empirical investigation, probably done by interdisciplinary teams of psychologists, philosophers, and scholars from law to show the actual importance of these desiderata for certain stakeholder classes.

Furthermore, in our overview, the desiderata's denotations stem (in most cases) directly from the source papers. However, some of these desiderata are closely related and, especially given the interdisciplinary research contributing to XAI, it is plausible that different authors actually mean to refer to the same desideratum but give it a different term or use the same term to refer to different desiderata. Consistent terminology and conceptual clarity for the desiderata are pivotal and there is a need to explicate the various desiderata more precisely. Different disciplines like law, philosophy, and psychology need to come together to discuss their conceptions of various desiderata to agree on common definitions of these desiderata. Without this, insights from different disciplines regarding the respective desiderata (e.g., what kind of explanatory information is required to satisfy a given desideratum) might not be adequately integrated into a common stream of research.

Additionally, we need research that more explicitly analyzes society's stakeholder classes affected by artificial systems. For instance, collaborating with sociologists could offer a broader or more nuanced picture of the classes of stakeholders that have to be considered within the scope of XAI. In any case, in order to comprehensively address the stakeholders' desiderata, we need a more detailed understanding of stakeholder classes and sub-classes. For this, it is promising to consult disciplines outside of computer science focusing on society (i.e., sociology) as well as individual differences within groups of society (i.e., psychology).

Furthermore, researchers from different disciplines may be able to take the perspective of certain stakeholder classes. By doing so, they can help to refine the list of desiderata. For instance, computer scientists can take the perspective of developers. Psychologists can take the perspective of users and affected people. Management scholars could take the perspective of deployers. Philosophers, political scientists, as well as researchers from law can take the perspective of regulators. 

Working together in interdisciplinary teams can, thus, contribute to a comprehensive consideration of important desiderata in a given context. However, comprehensiveness is just one side of the coin, the justification of desiderata is another. Concerning this justification, there are two main perspectives: one from ethics and one from jurisprudence. From an ethical perspective, we can judge whether a desideratum is compatible with some, many, or even all established moral theories. Similarly, legislation can be consulted to assess whether there are laws demanding (or prohibiting) to meet certain desiderata. When engaging in thorough moral and legal justification, we might conclude that there will be desiderata that are not justifiable. In high-stakes decisions, for instance, each individual user might want systems to do what is best for her. In the case of autonomous cars, drivers will probably want a car to decide in a way that makes it more likely that they will survive if the car faces an imminent accident \cite{Bonnefon2016}. There might be cases where such a decision is neither morally nor legally justifiable. In a less drastic example, users may ask for an explanation of why they received a low score on a personnel selection test. However, providing this explanation might render the given test obsolete because the explanation possibly enables participants to game the test \cite{Burrell2016}. We suggest that the given context, as well as moral and legal considerations are decisive factors when determining whether certain stakeholders' desiderata can be justified.

\section{Desiderata Satisfaction requires Understanding}
\label{sec:desrun}

In the previous section, we have introduced our claim that the need for XAI arises from stakeholders' desiderata. More precisely, the need arises in cases where certain stakeholders' desiderata are not (sufficiently) satisfied \cite{Arrieta2020, Preece2018, Felzmann2019, Mittelstadt2019, Floridi2018}. For this reason, we have to take a look at what it means for a desideratum to be satisfied.

\subsection{Facets of Desiderata Satisfaction}
\label{sec:facets}

We propose that the satisfaction of each desideratum can take two facets. We call these facets \emph{epistemic} and \emph{substantial} desiderata satisfaction, respectively. On the one hand, stakeholders want systems \emph{to have} certain properties that make them actually fair, transparent, or usable. In line with this, a desideratum (e.g., fairness) is substantially satisfied if a system sufficiently possesses the corresponding properties. On the other hand, stakeholders want \emph{to know} or be able \emph{to assess} whether a system (substantially) satisfies a certain desideratum (i.e., whether the system has the required properties). So, the epistemic facet of the fairness desideratum is satisfied for a stakeholder, if she is in a position to assess or know whether and to what extent the system is fair. Naturally, for XAI the epistemic facet is the most important one, since explanatory information can contribute to the satisfaction of the epistemic facet of every desideratum, whereas this is not the case for the substantial facet.

As an example, take the desideratum of having usable systems. A successful explanation process as depicted in our model may enable users to recognize whether a system is usable, and, optimally, also increase the system's usability to a certain degree. In this case, the epistemic satisfaction consists in the stakeholders being able to check whether a system or its outputs are usable for the task at hand. To a lesser extent, however, an explanation process can also contribute to the substantial satisfaction of the desideratum, since it provides additional knowledge about the system that makes it more usable for the stakeholder. For larger deficits in usability to be addressed, however, explanatory information might not directly help; for this, the entire artificial system may need to be redesigned.

Depending on the desideratum, the two facets are correlated to a certain degree (possibly even completely, when satisfying the epistemic facet also completely satisfies the substantial facet). To illustrate, consider the desideratum of retaining user autonomy in human-in-the-loop scenarios. Let us assume that an explanation process has helped to satisfy the epistemic facet of this desideratum to a certain degree, as it has enabled the user to assess the extent to which she can retain her autonomy in making decisions based on the recommendations of the system. Additionally, the more understanding a user has about a system's output, the more autonomous she can decide based on it. Thus, the explanation process has helped to satisfy the epistemic and the substantial facet of this desideratum. Hence, in this case, the substantial and the epistemic facet of desiderata satisfaction are highly correlated.

Now, consider the desideratum that systems adhere to certain ethical principles. When having sufficient information about a system, regulators can evaluate whether this system complies with ethical standards. Again, the explanation process serves to satisfy the epistemic facet of this desideratum. However, this does not directly make the system's processes and outputs more likely to comply with ethical standards. Consequently, explanation processes can at most indirectly satisfy the substantial facet of this desideratum: based on the understanding obtained by the explanation process, faults can be identified and steps to improve systems regarding their ethical properties can be initiated. In this case, the epistemic and the substantial facet of desiderata satisfaction are only loosely correlated. 

On the one hand, the distinction of these two facets shows that explanation processes can contribute to the satisfaction of all epistemic facets of desiderata concerning artificial systems. On the other hand, it shows that an explanation process alone does sometimes not suffice to satisfy the substantial facet of desiderata concerning artificial systems. In many cases, however, the epistemic satisfaction enables the substantial one. This means that even if a better understanding of the systems triggered by explanatory information does not always directly lead to the substantial satisfaction of the desiderata, it can form the necessary basis for achieving it. As the epistemic satisfaction of a desideratum is closely linked to a better understanding of a system, understanding is the pivotal point for all endeavors of satisfying desiderata.

\subsection{Understanding}

Throughout the history of XAI research, authors have highlighted the central importance of understanding in XAI (e.g., \cite{Buchanan1984, Clancey1983, Dhaliwal1996, Ribeiro2016, Koehl2019, Paez2019}). The overall goal of XAI is to advance human understanding of artificial systems in order to satisfy a given desideratum. There is an ongoing debate in the philosophical literature about what constitutes understanding \cite{DeRegt2017, Baumberger2017, Malfatti2019}, and a comprehensive review of this concept is beyond the scope of the current paper (see, for instance, \cite{Baumberger2017} for a review on understanding, \cite{Baumberger2014, Lambert1991} for papers on the concept of understanding, \cite{Keil2006, Lombrozo2006, Chi1994} for the relation between explanations and understanding, or \cite{Mayer1992} for the related topic of cognitive processes in knowledge acquisition; furthermore, see \cite{Mueller2019} for a broad overview on the theoretical basics of understanding relevant for XAI research). Some aspects of understanding, however, are typically agreed upon: There are different \emph{depths} and \emph{breadths} of understanding (in the following, we will use the term \emph{degree of understanding} to address depth and breadth of understanding, similar to \cite{Kelp2015, Baumberger2017}), and there are different \emph{kinds} of understanding \cite{Baumberger2014, Paez2019}. 

For the evaluation of explainability approaches it will, thus, be crucial to determine stakeholders' understanding of artificial systems. Examining understanding of software has a long history in human-computer interaction and education \cite{Zhu2018, Feltovich2001}. Typically, when referring to the understanding of an artificial system, these disciplines also refer to users' \emph{mental models} of systems. A mental model of a system can be understood as a mental representation of this system and its functioning. According to Rouse et al., mental models allow humans \enquote{to generate descriptions of system purpose and form, explanations of system functioning and observed system states, and predictions of future states} \cite{Rouse1986}. In other words, mental models allow humans to mentally simulate aspects of a system, for instance, in order to understand the causes of its decision-making \cite{Hoffman2018}.

Translated to the case of artificial systems, this means that we could examine understanding by investigating stakeholders' mental models of a system \cite{Zhu2018, Hoffman2018}: how well does a person's mental model mirror the actual system? Are there gaps in the current understanding of a system's functioning? Are people overconfident that they understand how a system works (when they actually only have an illusion of understanding) \cite{Rozenblit2002}? Are there learned misconceptions about systems and their outputs that need to be revised \cite{Kuhn2001}? For example, it is possible to investigate a stakeholder's mental model through think-aloud techniques where stakeholders are tasked to describe systems and their inner workings, or by letting stakeholders draw their mental model of a given system (for an overview of methods to elicit mental models, see \cite{Hoffman2018}). Similarly, it is possible to measure understanding by capturing what humans' mental models enable them to do in relation to a system and its outputs. For example, Kulesza et al.~\cite{Kulesza2013} used an \emph{explanation task} to assess whether participants understood what kinds of information are used to predict outcomes. Other studies used \emph{prediction tasks} to assess, for instance, whether participants can anticipate which predictive model would produce better outcomes \cite{Ribeiro2016}, whether participants can predict what outcome a predictive model would produce for a person with a given profile \cite{Cheng2019}, or whether participants can foresee the influence of a given feature on an outcome \cite{Cheng2019}. Another possibility would be to use \emph{manipulation tasks} in order to assess whether people understood what kind of information might add to the predictive accuracy of a model (e.g., like in \cite{Ribeiro2016}). Further possible tasks might be \emph{perception tasks} (e.g., naming of recognized characteristics of a model) or \emph{imagination tasks} (e.g., estimating what a model would predict for a given input), all of which would reflect different degrees of understanding. Furthermore, all of these tasks can reveal misconceptions of a system's functioning or knowledge gaps that need to be adjusted or filled with additional or alternative explanatory information. Additionally, what all of these ways to capture stakeholder understanding have in common is that they might help us to examine whether a given desideratum has the potential to be satisfied. For example, if a developer has understood a system in a way that she can imagine situations under which a system might fail, her ability to make the system more robust most likely increases.

Furthermore, there is an initial degree to which stakeholder understand artificial systems. Specifically, stakeholders without any prior experience with a given system will likely start with a degree of understanding that corresponds to their (background) knowledge of artificial systems or arose from initial instructions they have received regarding the system \cite{Hoffman2018, Tullio2007}. Thus, they will have an incomplete or even faulty mental model of the given system \cite{Paez2019}. For instance, a stakeholder might know (or might be informed) that machine-learning based systems are usually trained on historical data in order to predict new data. This degree of understanding can, then, be augmented (e.g., with explanatory information generated from explainability approaches). With a higher degree of understanding (and, consequently, a more detailed and accurate mental model of a system), a stakeholder might understand what kind of training data underlie a given system, what kind of algorithm is used for a given system, or what kind of output data a system produces \cite{Mitchell2019, Sokol2020, Hoffman2018}. Thus, with increasing degrees of understanding, stakeholders will be able to assess whether a given system has desired characteristics and adequate processes, or produces expected outcomes. In other words, an increasing degree of understanding will satisfy the epistemic facet of more and more desiderata. For the satisfaction of the desideratum's substantial facet, however, the opposite might sometimes be the case, as we will discuss in the next section.

\subsection{The Relation Between Understanding and Desiderata Satisfaction}

In certain cases, a stakeholder's degree of understanding and the extent of desiderata satisfaction are positively correlated. For instance, if the desideratum is to retain autonomy in interaction with a system, usually a higher degree of understanding satisfies the epistemic and substantial facets of this desideratum to a greater extent. However, there are also more complex cases. Assume that the desideratum is to trust a certain system. Acquiring a higher degree of understanding will increase a stakeholder's epistemic satisfaction of this desideratum (i.e. she can better assess whether and to what extent to trust the system), but the substantial facet (i.e., her actual trust) can be influenced in a negative way. When a stakeholder still possesses a low degree of understanding, she is likely to be unaware of problematic features a system has in certain contexts (e.g., in complex environments) or with certain kinds of input data (e.g., noisy inputs). So, with a low degree of understanding, a stakeholder is likely to trust a system (although inadequately) \cite{Kizilcec2016, Binns2018}. In contrast, with a higher degree of understanding, the stakeholder is able to recognize or even explain the conditions under which a system will tend to fail. Therefore, she is more aware of the system's problematic features, and this may, consequently, decrease her trust in it \cite{Cheng2019, Kizilcec2016}.

Additionally, it can happen that understanding contributes to the satisfaction of a single desideratum of a stakeholder to a greater extent, while the satisfaction of other desiderata for the same stakeholder suffers \cite{Sokol2020}. To illustrate, take the trade-off between transparency and non-gameability of systems. Deployers of systems want to comply with legislation and, consequently, want their systems to be transparent. Understanding is a necessary condition for perceived transparency. However, making systems more transparent can diminish another desideratum of deployers, the systems' non-gameability \cite{Burrell2016}. In other words, it should not be possible for particular users to manipulate a system in such a way that they can systematically evoke beneficial outputs for them. However, more transparency caused by a higher degree of understanding may enable some people to exploit the system \cite{Sokol2020}. In a personnel selection test, for example, a better understanding of the selection system may enable participants to game the test, preventing its proper use (i.e., selecting suitable applicants).

This points to another potential trade-off to be considered regarding the relation between an advanced understanding and desiderata satisfaction. Felzmann and colleagues \cite{Felzmann2019} argue that different stakeholders might hold different expectations regarding the extent to which a single desideratum has to be satisfied. It is furthermore possible that, while desiderata of one stakeholder are influenced positively by an increase in understanding, desiderata of other stakeholders suffer. An example of such a case was described by Langer et al.~\cite{Langer2018}. They provided additional information accompanying an automated personnel selection system for their participants, which resulted in more perceived transparency, but at the same time reduced acceptance of the system (for a similar finding see \cite{newman2020}). In such cases, it can happen that the two desiderata of transparency and acceptance arise from different perspectives and characteristics of stakeholders. For instance, legislation (i.e., a regulator) might call for transparency of systems, whereas a company using a system (i.e., a deployer) desires the system to be accepted. Explaining the system will, then, lead to the satisfaction of the legal desideratum, but at the price of impairing the company's desideratum.\footnote{Note that this example also describes a situation where two desiderata of the company are in conflict: user acceptance and adhering to legislation.}

These examples indicate that the degree and kind of understanding of artificial systems which explainability approaches should evoke may depend on trade-offs between a variety of desiderata from a variety of stakeholders. Consequently, the development, implementation, and use of explainability approaches should go hand in hand with a case-by-case evaluation of the relevant stakeholders' desiderata. While estimating the effects of explainability approaches, it is central to investigate the perspective of not only one, but all stakeholders who potentially have a variety of (conflicting) desiderata with regard to a given system.

\subsection{Factors Influencing the Relation Between Understanding and Desiderata Satisfaction}
\label{sec:conuds}

Characteristics of the context in which artificial systems operate moderate the relation between understanding and desiderata satisfaction and may affect the degree and kind of understanding necessary to satisfy a given desideratum. There is no agreed-upon definition of the term \enquote{context} \cite{Bazire2005} and a deeper dive into the discussion about the term goes beyond what we can achieve in this paper (for discussions on this topic, see \cite{Bazire2005, Dourish2004}). Following Dourish \cite{Dourish2004}, we hold that the context is set by a given situation, in the interaction between a stakeholder, an artificial system, a given activity or task, and an environment. This makes it impossible to anticipate all contextual influences that will affect the process of how explainability approaches aim to satisfy stakeholders' desiderata without knowing the concrete situation. Nevertheless, it is crucial to consider and anticipate these contextual influences when evaluating and developing explainability approaches.

For instance, what is at stake in a given situation can affect the relation between understanding and stakeholders' desiderata. That is, whether a context is a high or low stakes scenario may determine the degree of understanding necessary to satisfy a given desideratum. Research indicates that certain situations tend to require a greater understanding of an event than other situations. Specifically, situations where instrumental, relational, moral, or legal values are at stake might be more likely to require extensive understanding \cite{Bobocel2005, Folger2001, Shaw2003}. Instrumental values are at stake when there are (personal, economical etc.) benefits or losses to expect in a specific situation (e.g., when an artificial system handles financial transactions). Relational values are at stake when important interpersonal relationships might be affected by an event (e.g., when artificial systems are used for employee layoff). Moral values are at stake when moral rights might be violated (e.g., when using artificial systems for sentencing in court). Finally, legal values are at stake when legal rights might be violated (e.g., when an artificial systems outputs conflict with the right of non-discrimination). Depending on the concrete situation (e.g., being in an autonomous car), instrumental, relational, moral, and legal values determine the stakes of a situation. Additionally, identifying which of these values stakeholders regard as relevant or which values are indeed relevant in which situations may allow for drawing inferences about whether a situation is (considered) high or low stakes. Consequently, these values serve as an orientation for when stakeholders are more likely to demand higher degrees of understanding. Specifically, this means that for the satisfaction of a given desideratum in a low stakes scenario, a lower degree of understanding might be sufficient compared to high stakes scenarios. 

Furthermore, contexts involving artificial systems may differ in their \denquote{outcome favorability} \cite{Brockner1996, Wang2020}. In the case of an unfavorable outcome, people are more likely to call for additional information in order to understand the reasons for the outcome, and so to be able to better assess and control further similar outcomes \cite{Brockner1996, Gregor1999}. On the one hand, it may be that increasing the extent of understanding in the case of favorable outcomes has little potential to positively affect a desideratum's satisfaction. On the other hand, a better understanding can be central to positively affecting a desideratum's satisfaction in the case of a unfavorable outcome. For instance, perceived fairness is a central desideratum for many decision situations \cite{Lind2002}. While understanding may have negligible effects on perceived fairness under favorable outcomes, it can improve perceived fairness under unfavorable outcomes \cite{Colquitt2002}. Supporting this claim, Kizilcec \cite{Kizilcec2016} found that available explanatory information only affected perceived fairness when people's expectations concerning an outcome were violated. There may, however, be certain conditions under which a better understanding (negatively) affects perceived fairness even when people experience favorable outcomes \cite{Shaw2003}.

What is at stakes in a situation and outcome favorability are just two of many possible contextual influences on the relation between understanding and desiderata satisfaction. Further candidates are the application context (e.g., at home, at work), time constraints \cite{Liu2012}, and social circumstances (e.g., whether there are other people present in a given situation; \cite{Dourish2004}).

\subsection{Interdisciplinary Potential}
\label{sec:ipund}

The interdisciplinary potential we see in the relation between understanding and desiderata satisfaction is described in research questions such as: (a) How does stakeholders' initial degree of understanding or prior knowledge of artificial systems relate to their desiderata satisfaction? (b) What are the trade-offs between desiderata of a single stakeholder and/or desiderata of multiple stakeholders and what are the implications of understanding regarding these trade-offs? (c) How does the degree and kind of understanding relate to the satisfaction of desiderata? (d) How do task or context affect the relation between understanding and desiderata satisfaction in a given situation?

Scholars from educational sciences could collaborate with computer scientists in order to investigate how to design adequate instructions to achieve a proper basic understanding or background knowledge of artificial systems that can serve as a general basis to partially satisfy a large variety of desiderata. Furthermore, it is necessary to involve psychologists in order to experimentally examine the relation between understanding and desiderata satisfaction as well as influences affecting this relation. In this regard, it will be central to determine what it means for a certain desideratum to be satisfied. This requires finding an adequate way of measuring the satisfaction of this desideratum (e.g., using self-report measures, expert interviews, legal analyses), as well as understanding the requirements for it to be satisfied (e.g., defining minimum legal standards, enabling stakeholders to perform a specific task successfully). In practice, this involves having an elaborated research design, clear conceptual definitions, appropriate operationalization and measurement methods, and fitting research disciplines for iterative (empirical) research.

Finally, scholars from requirement engineering may help to understand relationships between several desiderata. Based on their results, scholars from law and from philosophy can help to determine which trade-offs are morally and/or legally justified. In general, an interdisciplinary collaboration can contribute to being aware of potential relationships and trade-offs between certain desiderata.  

\section{Understanding Requires Explanatory Information}
\label{sec:unrei}

Providing explanatory information of a given phenomenon is the default procedure to facilitate its understanding \cite{Wilkenfeld2014, Wilkenfeld2016, Lambert1991}. Explanatory information helps humans to navigate complex environments by facilitating better understanding, predictions, and control of situations \cite{Keil2006, Lombrozo2011}. Such information narrows down possible reasons for events, decreases uncertainty, corrects misconceptions, facilitates generalization and reasoning, and enables a person to draw causal conclusions \cite{Keil2006, Lombrozo2011, Williams2013, Rozenblit2002}.

In the context of XAI, explanatory information puts stakeholders in a position to grasp generalizable patterns underlying the production of a system's outcomes (e.g., LIME \cite{Ribeiro2016}). These patterns allow for drawing inferences about (potentially causal) connections between the system's inputs and outputs, or for narrowing down the possible ways in which the system might have failed. Additionally, these patterns may help stakeholders to tell apart correct but unexpected behavior from malfunctions in order to to debug the system. In general, all of this can reduce people's uncertainty concerning a system (e.g., uncertainty concerning how to behave towards it, how to react to it, what to think of it, whether to recommend it, or whether to disseminate it; \cite{Sokol2020}). Overall, explanatory information should lead to a better understanding, which should, in turn, positively affect the satisfaction of stakeholders' desiderata. Depending on the explanatory information, different degrees and kinds of understanding can be acquired \cite{Paez2019, Kelp2015}. For this reason, we have to shed more light on what characteristics of explanatory information are. 

\subsection{Characteristics of Explanatory Information}

Important characteristics of explanatory information concern its kind and its presentation format. There are various kinds of explanatory information: teleological (i.e., information that appeals to the function of the explanandum; \cite{Lombrozo2012}), nomological (i.e., information that refers to laws of nature; \cite{Hempel1965}), statistical relevance (i.e., information that is statistically relevant to the explanandum; \cite{Salmon1984, Gardenfors1988}), contrastive (i.e., information that highlights why event P happened and not event Q; \cite{Miller2019}), counterfactual (i.e., information that appeals to hypothetical cases in which things went differently; \cite{Wachter2018}), mechanistic (i.e., information that appeals to the mechanisms underlying a certain process; \cite{Craver2007}), causal (i.e., information that appeals to the causes of an event; \cite{Pearl2009, Spirtes2001}), network (i.e., information that appeals to the topological properties of a network model describing a system; \cite{Borsboom2018}), and many more. 

Concerning the presentation format, there are also various possibilities. Roughly, we can distinguish between text-based and multimedia presentation \cite{Gregor1999}. A text-based presentation can be a natural language text, a rule extracted from a rule-based system, an execution trace, or simply the program's source code. A multimedia presentation can include graphics, visualizations, images, and animations. For instance, heat-maps of neural activity are a popular presentation format for explanatory information of neural networks \cite{Arrieta2020}.

Aside from the very general characteristics of kinds and presentation format, there are further characteristics of explanatory information that influence how and whether it evokes understanding: soundness (i.e., how accurate the information is), completeness, novelty (i.e., whether the information is new for the recipient), and complexity (e.g., depending on the number of features and on the interrelation between the features the information contains; \cite{Liu2012}) are just some of the various examples of further characteristics of explanatory information (for more, see \cite{Sokol2020}). 

The importance of acknowledging the characteristics of explanatory information is highlighted by research in cognitive and educational psychology that shows that effects of explanatory information can vary depending on their characteristics \cite{Keil2006, Mayer1992, Shaw2003}. Take complexity as an example: studies show that people prefer simpler information (e.g., information mentioning fewer causes; \cite{Lombrozo2007}). Another example is the finding that explanatory information that aligns with a stakeholder’s goals in a certain situation is preferred \cite{Vasilyeva2017}. Vasilyeva et al. \cite{Vasilyeva2017} showed that people evaluate teleological explanatory information compared to mechanistic explanatory information as more useful when asked to name the function of a phenomenon, and, conversely, perceive mechanistic explanatory information as more useful when asked to name the cause of an event.

\subsection{Factors Influencing the Relation Between Explanatory Information and Understanding}
\label{sec:conexu}

The relation between explanatory information and understanding is influenced by a variety of characteristics of the stakeholders, the context, and interactions between these characteristics \cite{Bellotti2001, Bobocel2005, Shaw2003}. Considering these influences is central to evaluate and develop explainability approaches.

\paragraph{Characteristics of Stakeholders}
Since every stakeholder possesses an individual degree of understanding of a system, an individual ability to understand, and an individual set of desiderata that are or need to be satisfied to a certain extent, the characteristics of stakeholders who receive explanatory information influence the relation between explanatory information and understanding \cite{Preece2018}. Some of the characteristics that most obviously influence this relation are the stakeholders' background knowledge, beliefs, learning capacities, and desiderata they have concerning respective systems \cite{Hartley2001, Kauffman2015, McNamara1996, Sokol2020}. For instance, explanatory information including technical details might increase an expert developer's degree of understanding while technical details can hamper understanding for novice developers or other (non-expert) stakeholders \cite{Williams2013}.

Furthermore, desiderata that are salient for a respective stakeholder can influence the relation between explanatory information and understanding. Specifically, as stakeholders engage with information in order to advance their understanding of artificial systems, their motivation and prior beliefs may affect how they interpret a given set of information \cite{Wang2019}. For instance, if a stakeholder's primary desideratum is to ensure that a system provides fair outputs, they will scrutinize explanatory information for signs of bias that might lead to unfair outcomes. In contrast, if a stakeholder's primary desideratum is to improve a system's predictive accuracy, they might pay special attention to information providing insights on how to improve system performance. This indicates that, depending on the given desideratum, the same amount and kind of information can lead to different degrees and kinds of understanding. Consequently, it is important to provide the appropriate information for the given purpose. These assumptions are supported by research proposing that human reasoning processes can be strongly influenced by explanatory information. Such information has the potential to improve human decision-making but may also hamper it (e.g., when explanatory information attenuates human biases or when humans need to invest too much effort to use the information \cite{Wang2019, Gregor1999}). 

The list of individual differences influencing the relation between explanatory information and understanding goes beyond the scope of this article and covers personality traits such as conscientiousness \cite{Goldberg1981}, need for cognition\footnote{Need for cognition is a personality trait that distinguishes people who like to put effort into cognitive activities from those who prefer less cognitively demanding activities.} \cite{Cacioppo1982, Haugtvedt1992}, and need for closure \cite{DeBacker2009, Webster1994}, as well as people's preferences for detailed versus coarse explanations \cite{Fernbach2013}, or age differences between stakeholders \cite{Hasher1988}. 

\paragraph{Characteristics of the Context}
Time pressure can be a relevant contextual influence \cite{Liu2012, Sokol2020}. For a stakeholder under high time pressure, the same explanatory information may lead to less understanding as compared to situations where she is under low time pressure \cite{Ackerman2012}. Workload is another contextual influence \cite{Prewett2010}. In situations of high perceived workload, the same explanatory information can lead to a different degree of understanding compared to low workload conditions. Similar things are true for situations where it is more likely that stakeholders will experience higher levels of stress \cite{Starcke2008, Lupien2007} (e.g., high stake and high risk situations, multitasking environments; \cite{Hancock2007, Liu2012}).

In general, depending on the situation and task at hand, the effects of explanatory information on understanding may differ. Therefore, it is important to investigate the contextual conditions of a stakeholder's interaction with an artificial system in detail when theorizing about how explanatory information can best improve understanding. Given the impact of the context and given the fact that it might not be possible to anticipate all relevant contextual influences \cite{Dourish2004}, it is especially important to assess the effects of explanatory information in laboratory and in field settings \cite{DoshiVelez2017, Sokol2020}. Although it has advantages to investigate the effects of, for instance, stress on the relation between explanatory information and understanding in controlled laboratory settings, such results may not translate to field settings. This finding is in line with calls for experiments involving human participants in proxy tasks in order to show that given explanatory information not only elicits understanding when simulating a context, but also under real world conditions (that is, to investigate whether given explanatory information is equally valuable in the wild) \cite{DoshiVelez2017, Sokol2020}. Although it will be impossible to fully anticipate how the context will alter the relation between explanatory information and understanding (e.g., because the interpretation of contextual influences depends on the relevant stakeholders \cite{Bazire2005}), at least it is crucial to be aware that contextual influences may be central for the success or failure of explanatory information. 

\paragraph{Interactions Between Characteristics of Explanatory Information, Stakeholders, and the Context}

Finally, characteristics of explanatory information, stakeholders, and the context can interact in ways that affect how stakeholders engage with explanatory information to advance their understanding of artificial systems. For instance, the level of detail of explanatory information can interact with the prior knowledge of a stakeholder. Whereas expert developers' understanding may benefit to a higher degree from explanatory information with much detail, this level of detail can have the contrary effect for novice developers \cite{Shaw2003, Chazette2019, Chazette2020, Gregor1999}. At the same time, when novice users want to learn how to use a system, they might want detailed explanatory information whereas when expert users want to use the system for task fulfillment, every unnecessary piece of information could lead to the rejection of the system \cite{Gregor1999}. Further, if the context changes, the relations between explanatory information and understanding are prone to change as well. For instance, as soon as there is time pressure in the aforementioned situations, it is plausible that neither expert nor novice developers or users pragmatically benefit from too detailed explanatory information.

\subsection{Interdisciplinary Potential}

The following research questions reinforce calls for extensive validation and experimental studies investigating the effects of explainability approaches in different contexts and in relation to different stakeholders \cite{DoshiVelez2017, Sokol2020} (we refer readers to Sokol et al.~\cite{Sokol2020} for a description of further important characteristics of explanations, stakeholders, and contexts that affect the relations between explanations and stakeholders' understanding): (a) How should we classify explanatory information? (b) How does different explanatory information lead to understanding? (c) How do different stakeholders engage with explanatory information? (d) How can we optimally evoke understanding through explanatory information? (e) How do stakeholder differences (e.g., background knowledge, personality characteristics) affect understanding? (f) How do contextual influences (e.g., different levels of risk, multi-tasking environments, time pressure) affect the relation between explanatory information and understanding? (g) How should explanatory information be designed? What should it include? What kind of presentation format (e.g., textual, graphical) is appropriate? How can stakeholders interactively engage with this information?

On the one hand, these research questions call for a unified classification of explanatory information. Often, computer scientist classify explanatory information based on its presentation format or based on the explainability approach it originated from (e.g., \cite{Arya2019}). Philosophers and psychologists, however, usually classify explanatory information in terms of the kinds mentioned above (e.g., causal or nomological; \cite{Woodward2019}). In order to prevent the debate from drifting apart, philosophers, psychologists, and computer scientists should collaborate to find standardized ways to classify explanatory information. On the other hand, these research questions call for empirical evaluation of the value of different kinds of explanatory information for different stakeholders, under a variety of contexts and under the consideration of different desiderata. This may primarily call for empirically working psychologists, as they have the tools and expertise to design experimental studies, for philosophers to determine the qualities of good explanatory information, and for computer scientists to adjust the presentation of explanatory information as well as possible. Furthermore, cognitive scientists might need to examine what exactly understanding of artificial systems actually means and how to measure it (see \cite{Hoffman2018} for ideas of how to capture human understanding of artificial systems).

However, deriving explanatory information of artificial systems is a task that in itself requires a lot of interdisciplinary research. The following section completes the specification of the concepts and relations in our model by describing that, in order to get the required explanatory information from the systems, there is the need for fitting explainability approaches.

\section{Explanatory Information Requires Explainability Approaches}
\label{sec:eirea}

In order to provide explanatory information that facilitates understanding and, thus, affects the satisfaction of the desiderata of the different classes of stakeholders, XAI research has developed a wide variety of explainability approaches. These approaches encompass methods, procedures, and strategies that provide explanatory information to help stakeholders better understand artificial systems, their inner workings, or their outputs. A specific explainability approach is characterized by all steps and efforts that are undertaken to extract explanatory information from a system and to adequately provide it to the stakeholders in a given context. Explainability approaches can take many guises and the literature commonly distinguishes two families of approaches (e.g., \cite{Sokol2020, Hall2019, Guidotti2019, Arrieta2020, Arya2019, Adadi2018}): \emph{ante-hoc} and \emph{post-hoc} approaches. 

\subsection{Families of Explainability Approaches}

Ante-hoc approaches aim at designing systems that are inherently transparent and explainable. They rely on systems being constructed on the basis of models that do not require additional procedures to extract meaningful information about their inner workings or their outputs. For example, decision-trees, rule-based models, and linear approximations are commonly seen as inherently explainable (given they have a limited size) \cite{Guidotti2019, Mittelstadt2019}. A human can, in principle, directly extract information from these models in order to enhance her understanding of how the system works or of how the system arrived at a particular output. Unfortunately, this way of deriving explanatory information from transparent models might only be useful for stakeholders with a certain expertise \cite{Miller2017, Gilpin2018b}. For this reason, ante-hoc approaches make systems directly explainable only for developers and members of other stakeholder classes that possess enough expertise about artificial systems. Furthermore, ante-hoc explainability can also lead to a loss of predictive power \cite{Arrieta2020, Felzmann2019} and not all systems can be designed inherently explainable.

Post-hoc approaches try to circumvent the aforementioned shortcomings. Such approaches are, in principle, applicable to all kinds of models. The difference to ante-hoc approaches is that post-hoc approaches do not aim at the design-process of a particular system, but at procedures and methods that allow for extracting explanatory information from a system's underlying model, which is usually not inherently transparent or explainable in the first place \cite{Lipton2018, Guidotti2019, Hall2019, Sokol2020}. Post-hoc approaches are, for example, based on input-output analyses or the approximation of opaque models by models that are inherently explainable. 

In many cases, however, post-hoc approaches are restricted with respect to how they present explanatory information. That is, given a specific model or one of its outputs, the information an approach will provide on repeated usage (and the format in which the approach provides the information) will be similar. Hence, for post-hoc approaches the same holds as for ante-hoc approaches: it is not guaranteed that all stakeholders are able to understand the provided information in the given format \cite{Miller2017, Gilpin2018b}. So, the explanatory information accessible from both, post-hoc and ante-hoc approaches is often only interpretable for developers or other expert stakeholders. This means that this information does not directly facilitate understanding for non-experts \cite{Gilpin2018b}.

One solution to this is to combine several explainability approaches in order to cover a broad range of different information and presentation modes. Another solution that has received increasing attention is to have recourse on interactive explainability approaches \cite{Adadi2018, Wang2019}. Interactive approaches are based on the idea that the user or some other stakeholder is provided with more in-depth information concerning a system if the information she initially received does not suffice. Based on her needs, the person interacting with the system can call for information about specific aspects of a decision or request a presentation in a different format. To date, however, approaches that are fully interactive remain rare \cite{Arya2019, Adadi2018}.

A third solution is to have a human facilitator (e.g., an expert stakeholder) explain a system to other stakeholders. For instance, when regulators want to satisfy their desiderata concerning artificial systems, there will be cases where they do not directly interact with an artificial system. Instead, a human facilitator (e.g., an expert user or a developer) will do so and derive suitable explanatory information for the regulator. In a sense, this process introduces a desiderata hierarchy based on the stakeholders into Figure \ref{fig:stakeholder}. In other words, one desideratum might have to be satisfied for one stakeholder class before some other desideratum (for another stakeholder class) can be satisfied. For example, the facilitator's desideratum (in this case: to be able to explain a system to a regulator), must be satisfied first as a precondition for the satisfaction of a regulator's desideratum (e.g., to be able to assess the fairness of a system). Thus, we have the complete processes in Figure \ref{fig:ourxai} nested within the explainability approach. Having a human facilitator (e.g., a developer) deriving explanatory information (assisted by an explainability approach) for another stakeholder (e.g., a regulator) can be considered a hybrid human-system approach to explainability.

\subsection{Factors Influencing the Relation between Explanatory Information and Explainability Approaches}

There are further characteristics of explainability approaches that are worth mentioning, since they are likely to influence the provided kind of explanatory information or its presentation format. First, it is important to distinguish post-hoc approaches that work regardless of the underlying model type (so-called \emph{model-agnostic} approaches) from ones that only work for specific (types of) models (so-called \emph{model-specific} approaches). Model-agnostic approaches aim to deliver explanatory information about a system solely by observing input/output pairs \cite{Ribeiro2016, Sokol2020, Guidotti2019, Hall2019}. Model-specific approaches do so while also factoring in specific features of the model at hand (e.g., by creating prototype vectors in a support vector machine) \cite{Sokol2020, Guidotti2019, Hall2019}. Model-agnostic approaches have the advantage of working for all types of models, but have the drawback that they tend to be less efficient, less accurate as well as less explanatory powerful (i.e., the explanatory information's level of detail is lower with regard to individual phenomena) than the former.

Second, previous research distinguished the scope of an explainability approach. Some approaches provide information about only single predictions of the model. The scope of these approaches is \emph{local} \cite{Sokol2020, Guidotti2019, Ribeiro2016, Hall2019}. Often, they offer visualized prototype outcome examples (e.g., \cite{Kim2014, Kim2016}). The more general type of approaches has a \emph{global} scope \cite{Sokol2020, Guidotti2019, Hall2019}. These approaches are designed to uncover the overall decision processes in the model. Here, the usual way to provide this information is by approximating complex models with simpler ones that are inherently explainable.

Depending on the explainability approach, the explanatory information provided will differ. Global explainability approaches, for instance, are likely to produce more complex information that requires more background knowledge by stakeholders to be understood. Local explanability approaches, on the other hand, only show a limited picture of a system's inner working and may not be representative of its overall decision-making processes. Based on these differences we can conclude that certain explainabilty approaches are more suitable for the satisfaction of given desiderata of specific stakeholder classes than others.

To elaborate, take users who want to calibrate their trust in a system. They will need a different kind of information compared to users who want to have usable systems. In the first case, the explainability approach will likely need to extract information about the robustness of a system, about conditions under which outputs of the system are trustworthy and situations where users have to be aware that outputs might be misleading. In the case of usable systems, users might want information that directly matches their specific goals in a given task. Another example: if a desideratum of users is learning how to use a system, they may need a different kind of explanatory information compared to when they simply want to fulfill tasks with the help of an artificial system \cite{Carroll1987}. This is because users who want to learn how to use a system need more details whereas users who want to fulfill tasks need directly useful information in order not to reduce their productivity through overly detailed information. Differences in the information needed can also be due to differences in the perspective of stakeholders. Take the desideratum of fairness. Affected parties will more likely focus on aspects of individual fairness which may call for explainability approaches that facilitate local explainability. Regulators, however, may focus on more general notions of fairness calling for explainability approaches facilitating global explainability. 

\subsection{Interdisciplinary Potential}

The design of explainability approaches and the goal of providing adequate explanatory information with the potential to affect stakeholders desiderata, again, hold untapped interdisciplinary potential, with research questions such as: (a) How can we pinpoint what explanatory information an explainability approach should provide in which case? (b) How can we design interactive explainability approaches? (c) How can explainability approaches involving human facilitators be optimally designed? (d) How can we guarantee that an explainability approach provides the required explanatory information? (e) How should stakeholders' desiderata and their degrees of understanding be taken into account when generating explanatory information or when developing new explainability approaches?

For all of these research questions, we see a potential for collaboration between computer scientists, philosophers, psychologists, and cognitive scientists. For instance, philosophers and psychologist have to determine which information is needed to assess whether a system is fair, whereas computer scientists develop explainability approaches that provide this information, are aware of trade-offs between different approaches as well as of technical constraints. Furthermore, investigating how the examination of stakeholders and their desiderata narrows down the options of possibly successful explainability approaches might be a fruitful area for future research.

A more thorough discussion about which desiderata of what stakeholder class call for what kind of explainability approach goes far beyond what a single paper can achieve. However, in Section \ref{sec:use}, we will show that our model could inspire work that is necessary to evaluate the usefulness of an explainability approach in relation to a given desideratum. Furthermore, we will show that the model supports the development of new explainability approaches to satisfy a given set of desiderata. Thus, the next section aims to show how our model can lead to actionable insights for XAI research by analyzing the concepts and relations that we propose in Figure \ref{fig:ourxai}. By means of hypothetical application scenarios, we want to stimulate ideas about specific applications of our model.

\section{Bringing It All Together: Hypothetical Application Scenarios}
\label{sec:use}

Following, we will present how the previous sections come together within hypothetical application scenarios. These scenarios highlight the importance of different stakeholder classes and their desiderata. Furthermore, these scenarios emphasize that understanding affects the satisfaction of these desiderata, that explanatory information provided by explainability approaches facilitate understanding, and that analyzing and investigating these concepts and their relations is central for the aims of XAI as well as for the development of explainability approaches that can successfully satisfy stakeholder desiderata.

The main application of our model is derived from the idea that if an explanation process does not change a certain desideratum's extent of satisfaction, the corresponding explainability approach might not be a suitable means for satisfying the desideratum in the given context. Such a discovery (e.g., resulting from stakeholder feedback or from empirical investigation) can provide feedback regarding which explainability approaches and what kinds of explanatory information work for which desiderata in which contexts. Thus, this feedback can serve as an input for the improvement of explanation processes and, consequently, helps to evaluate, adjust, choose, and develop explainability approaches for a given purpose and context. 

We propose that each step in our model (Figure \ref{fig:ourxai}: Explainability Approaches $\rightarrow$ Explanatory Information $\rightarrow$ Understanding $\rightarrow$ Desiderata Satisfaction) allows for drawing inferences about the explanation process involving explainability approach(es), kind(s) of explanatory information, and stakeholder understanding. The following questions, which arise at different points in our model, are of particular interest:
\begin{itemize}
\setlength\itemsep{-0.5ex}
    \item Who are the relevant stakeholders and what are their specific characteristics? Which are the relevant desiderata in a specific context and are they satisfied? 
    \item Have the stakeholders acquired a sufficient degree and the right kind of understanding that allows for assessing whether given desiderata are satisfied and to facilitate their satisfaction? 
    \item Does the provided explanatory information and its format of presentation facilitate stakeholders' understanding in a given context and in consideration of the stakeholder characteristics? 
    \item Is the explainability approach able to provide the right kind and amount of explanatory information in the right presentation format? 
    \item Are there contextual influences hindering or promoting the satisfaction of desiderata through the explanation process? 
\end{itemize}
Investigating these questions requires empirical research, hypothesis testing and interdisciplinary cooperation, but should, eventually, aid in evaluating, adjusting, choosing, and developing explainability approach(es) and finding explanatory information in order to adequately satisfy stakeholders' desiderata.

With this in mind, we believe that our model is useful for several important application scenarios. First, our model is useful for choosing adequate explainability approaches for novel application contexts of artificial systems and for guiding the development of new explainability approaches. Specifically, our model can be used to inform projects on how to develop explainability approaches to satisfy certain desiderata for a given class of stakeholders. Second, our model is useful for evaluating why and at which stage an explainability process failed to contribute to satisfying the relevant desiderata. Let us assume that the use of an explainability approach does not lead to the satisfaction of a certain desideratum. Why is this the case? Is the explainability approach not suitable for the satisfaction of the desideratum (e.g., the explainability approach provides the false kind of explanatory information) and should be replaced by another approach? Or does the error lie somewhere else in the explanation process? In some cases, we may be able to adjust the explainability approach appropriately to achieve its intended purpose. 

\subsection{General Application Scenarios}

For structured attempts to evaluate, adjust, choose, or develop explainability approaches for a given context we roughly distinguish two scenarios, which we call the \emph{evaluation} and the \emph{discovery} scenario. In the evaluation scenario, we want to investigate whether the use of a specific explainability approach was adequate, and if not, what is needed to fix its shortcomings. In the discovery scenario, we want to find an adequate explainability approach to satisfy stakeholders' desiderata. This can take one of two forms: choosing among existing approaches or, if no adequate approach is available, developing a new one.

\emph{Stakeholder and Desiderata}. Both evaluation and discovery scenarios start with examining the stakeholders and clarifying their desiderata in the given context. In the discovery scenario, we have to examine what the relevant classes of stakeholders are and which desiderata they have concerning the application of a system under consideration. In the evaluation scenario, we have to check -- even in scenarios where already identified desiderata are satisfied -- whether the explanation process fits all relevant classes of stakeholders and all of their desiderata (and ensure that we did not overlook important stakeholders or desiderata).

For this, we need input from a wide variety of disciplines including but not limited to scholars from law, sociology, psychology, philosophy, and computer science. Such a combination of expertise and perspectives helps to identify relevant stakeholder classes and list their desiderata pertaining to a given context. Defining these desiderata falls within the expertise of philosophers. The elaboration of the desiderata's relevant moral and legal aspects is a task for ethicists and scholars from law. Furthermore, assessing contextual peculiarities and stakeholder characteristics relevant in a given context will require psychologists (on the individual level), sociologist (on the group level), as well as domain experts such as judges or personnel managers (for particular application scenarios).

\emph{Desiderata Satisfaction}. For the discovery scenario, the next step is to determine what it means that stakeholders' desiderata are satisfied in a specific context: to make estimates, to provide guidelines, and to create measures for when the identified desiderata are satisfied. For the evaluation scenario, we have to check whether the relevant desiderata are satisfied or not. If they are, the explanation process was successful and the chosen explainability approach appropriate. If they are not substantially satisfied, there are two possible cases. First, the necessary understanding was not acquired and, thus, the desideratum is also not epistemically satisfied. Then, an improvement of the stakeholders' understanding is required. Second, an adequate degree and kind of understanding is reached and, thus, the epistemic facet of the desideratum is satisfied. In this case, we may conclude that the regarded desideratum is not directly substantially satisfiable by means of explainability approaches. Regardless of the case, at this point we have to move on to investigate stakeholders' understanding.

Determining conditions for desiderata satisfaction will be an interdisciplinary task for psychologists, philosophers, and scholars from law. Furthermore, computer scientists and domain experts can give practical input on satisfaction conditions for the desiderata in their specific domain. Making the satisfaction of these desiderata measurable and examining their extent of satisfaction will be a job for psychologists who develop measures or tasks that help to assess the extent of desiderata satisfaction. Additionally, it can be a task for scholars from law or philosophers to provide clear guidelines for when a desideratum is satisfied.

\emph{Stakeholders' Understanding}. The discovery scenario continues by investigating and defining requirements for the stakeholders' understanding needed to satisfy the desiderata under consideration. Specifically, we have to determine the appropriate degree and kind of understanding concerning the system and its output that promise to enable the epistemic facet of desiderata satisfaction. At the same time, this means that we need to assess (e.g., in studies using tasks to measure stakeholders' mental models of a system, or in studies using tasks to reveal whether stakeholders were able to, for instance, explain a systems' functioning or predict a systems' behavior and outcomes) stakeholders' actual degree and kind of understanding. This is especially important in the evaluation scenario when deficits in the substantial desiderata satisfaction become apparent. This results from the circumstance that assessing whether the required degree and kind of understanding that has been achieved allows for drawing inferences as to whether there is a fundamental gap between the explanatory process and the substantial desiderata satisfaction, or whether the provided information is not appropriate to evoke understanding. This is due to the fact that the explanation process can only serve to maximize understanding. If a desideratum still remains (substantially) unsatisfied, its full satisfaction goes beyond the scope of any explainability approach.

Philosophers can help to investigate and explicate what it means to have a certain degree or kind of understanding. Building on this, psychologists can design ways to empirically assess and measure such understanding (e.g., is it necessary for a given desideratum that stakeholders are able to predict a system's outputs? Is it necessary that stakeholders are able to anticipate situations when systems will likely fail?). Furthermore, scholars from law can contribute conditions for traceability and auditability. 

\emph{Explanatory Information}. The next step in the discovery scenario is to pin down what explanatory information has the potential to facilitate the right kind and degree of understanding in a predetermined context. This implies an evaluation of different dimensions of explanatory information with respect to the expected effects within an explanation process. To do so, we have to pay attention to, for instance, the kind of information, its presentation format \cite{Gregor1999}, its quality \cite{Hoffman2018}, its amount \cite{Liu2012}, its completeness, its complexity, or its adequacy for the given context \cite{Liu2012}. In the evaluation scenario, we have to check whether an explainability approach provides explanatory information that sufficiently meets the previously identified requirements. Sometimes there will also be a need to re-evaluate whether the requirements concerning the information are indeed adequate.

In this respect, philosophers and other explanation scientists can help to distinguish between different kinds and features of explanatory information \cite{Mittelstadt2019, Miller2019}. Furthermore, scholars from law can examine current legislation to find out whether it prescribes certain kinds of explanatory information. In the case of the GDPR, for instance, they have to specify what it means to \enquote{provide [...] meaningful information about the logic involved} (GDPR Art.~13 (2)(f); \cite{Goodman2017}). Finally (and based on this differentiation), educational or cognitive psychologist have the task to characterize the explanatory information that is best suited to facilitate the required kind and degree of understanding for a certain context. 

\emph{Explainability Approach}. Insights from the assessment of the concepts and relations in our model can guide and inform the requirements for explainability approaches that aim to satisfy given desiderata. This is of particular importance for the discovery scenario, where the primary objective is to identify which approach is expected to be most appropriate for providing specific information. Assume that, in the evaluation scenario we are at a point where the desiderata are not satisfied, the adequate degree and kind of understanding is not evoked, and the required explanatory information is not delivered. In this case, it is necessary to investigate whether the explainability approach is even capable of producing explanatory information with the right features at all. All the insights that are available at this point can indicate whether an existing explainability approach provides explanatory information that is sufficient to satisfy stakeholders' desiderata.

Additionally, we can learn whether a given explainability approach has the potential to derive explanatory information that can satisfy stakeholders' desiderata to a certain degree, whether it is necessary to adjust the explainability approach, whether it is sufficient to choose another one, or whether we need to develop an entirely new one. At this stage, computer scientists who can improve, adjust, and design explainability approaches are the main contributors integrating the aforementioned insights. First, they have the abilities to assess what is technically feasible and possible. Second, they can actually implement the demands regarding the explainability approach resulting from the previous steps.

\subsection{Specific Application Scenario}

We will conclude our thoughts about the application of our model by means of a specific example. Consider a situation where users want a system that produces fair outputs (substantial facet). For this, we first need an explanation process that enables them to assess whether the system produces fair outputs (epistemic facet) and, consequently, we need to find an adequate explainability approach. We first clarify the relevant user sub-classes and their prototypical characteristics. Will it be, for instance, novice or expert users? Then, we determine whether other stakeholder classes also have to be considered. For example, do we need to consider the perspectives of regulators regarding fairness? Furthermore, we have to anticipate the context. Are we talking about a personnel selection tasks or court cases with completely different contextual peculiarities?

Subsequently, we determine what users mean when they desire fair outputs (i.e., we clarify the satisfaction conditions of the desideratum's substantial facet). What kind of algorithmic fairness do they expect? At the same time, it is important to be aware of other relevant desiderata as the satisfaction of other user desiderata could be affected when trying to satisfy the fairness desideratum. Similarly, there may be unanticipated effects on the stakeholders' desiderata. For instance, will using a specific explainability approach that proves suitable to assess system fairness also affect predictive accuracy of the artificial system? 

Next, we consider the desideratum's epistemic facet. Under which circumstances would users be enabled to assess whether the outputs of a system are fair? What do they need to understand with regard to a system's functioning or outputs? To answer these questions, we must determine what degree and what kind of understanding is appropriate for the epistemic satisfaction of the fairness desideratum and we need a detailed investigation of how contextual influences moderate this relation. Furthermore, we must be aware of given stakeholder characteristics (e.g., users' background knowledge).

When we have estimated what degree and kind of understanding is required to enable users to assess whether a system's outputs are fair, we can determine what kind of explanatory information facilitates this understanding. For example, we might need explanatory information about the influence of features based on protected attributes (e.g., race) on the system's outputs. Alternatively, we might need information that contrasts the treatment of different groups of people (e.g., minorities and majorities). With this knowledge, we can determine what explainability approach provides this kind of explanatory information. Most likely, it would be post-hoc, local approaches that provide explanatory information that either highlights feature relevance or that allow users to compare subsets of instances regarding their predicted outcomes. As a result, we have specific demands regarding the explainability approach for satisfying the fairness desideratum in the case under consideration.

With this knowledge we can either choose an adequate explainability approach from existing ones or design a new one. Afterwards, we can investigate whether the explanatory information resulting from the respective explainability approach leads to a better understanding of the system and its outputs. This means that we are now in a position to empirically evaluate the success of the selected approach and the corresponding explanation process. 

For this, we may conduct a stakeholder study and find that, given our definitions of desiderata satisfaction, the epistemic facet may be satisfied (e.g., users can actually and justifiedly assess whether the system produces fair outputs; users can explain whether and why the respective system's outputs are fair; users can predict what kind of inputs will lead to fair or unfair outcomes) but not the substantial one (i.e., the system's outputs are actually not fair). Then, we may have to conclude that the substantial facet of the desideratum is not satisfiable by altering the explainability approach. Nevertheless, satisfying the epistemic facet of the desideratum can help us to figure out how to satisfy the substantial facet of the desideratum beyond XAI-related strategies. For instance, we may obtain information that a developer can use to improve system fairness. In this case, the artificial system has to be adjusted or changed to additionally satisfy the substantial facet of desiderata satisfaction. 

On the other hand, if the stakeholder study reveals that the epistemic facet of the desideratum is not satisfied (e.g., users fail in explaining whether and why the respective system's outputs are fair), we can conclude that the users' degree or kind of understanding of the system and its outputs does not suffice. In many situations this may be the case because the explanatory information was not suitable to evoke the necessary understanding (e.g., the explainability approach did not provide explanatory information suitable to allow assessing whether the system produces fair outcomes). Depending on the context, the task, and stakeholders' individual characteristics, it may be necessary to iteratively adjust the characteristics of the explanatory information so that stakeholders engaging with this information achieve the right kind and degree of understanding. Alternatively, it may be necessary to adjust the explainability approach if it is not suitable to provide explanatory information useful for facilitating stakeholder understanding. 

\section{Conclusion}
\label{sec:conc}

With increasing numbers of people affected by artificial systems, the number of stakeholders' desiderata will continue to grow. Although the focus of XAI research has shifted towards human stakeholders and the evaluation of explainability approaches, this shift still needs to incorporate a comprehensive view of all stakeholders and their desiderata when artificial systems are used in socially relevant contexts as well as empirical investigation of explainability approaches with respect to desiderata satisfaction. This engenders an even more pressing need for interdisciplinary collaboration in order to consider all stakeholders' perspectives and to empirically evaluate and optimally design explainability approaches to satisfy stakeholders' desiderata. The current paper has introduced a model highlighting the central concepts and their relations along which explainability approaches aim to satisfy stakeholders' desiderata. We hope that this model inspires and guides future interdisciplinary evaluation and development of explainability approaches and, thereby, further advances XAI research concerning the satisfaction of stakeholders' desiderata.

\bibliography{mybibfile}

\begin{thebibliography}{100}
\expandafter\ifx\csname url\endcsname\relax
  \def\url#1{\texttt{#1}}\fi
\expandafter\ifx\csname urlprefix\endcsname\relax\def\urlprefix{URL }\fi
\expandafter\ifx\csname href\endcsname\relax
  \def\href#1#2{#2} \def\path#1{#1}\fi

\bibitem{Brock2018}
D.~C. Brock, Learning from artificial intelligence's previous awakenings: The
  history of expert systems, {AI} Magazine 39~(3) (2018) 3--15.
\newblock \href {http://dx.doi.org/10.1609/aimag.v39i3.2809}
  {\path{doi:10.1609/aimag.v39i3.2809}}.

\bibitem{Clancey1983}
W.~J. Clancey, The epistemology of a rule-based expert system -- {A} framework
  for explanation, Artificial Intelligence 20~(3) (1983) 215--251.
\newblock \href {http://dx.doi.org/10.1016/0004-3702(83)90008-5}
  {\path{doi:10.1016/0004-3702(83)90008-5}}.

\bibitem{Swartout1983}
W.~R. Swartout, Xplain: A system for creating and explaining expert consulting
  programs, Artificial Intelligence 21~(3) (1983) 285--325.
\newblock \href {http://dx.doi.org/10.1016/S0004-3702(83)80014-9}
  {\path{doi:10.1016/S0004-3702(83)80014-9}}.

\bibitem{Johnson1993}
H.~Johnson, P.~Johnson, Explanation facilities and interactive systems, in:
  Proceedings of the 1st International Conference on Intelligent User
  Interfaces ({IUI}), Association for Computing Machinery, New York, NY, USA,
  1993, pp. 159--166.
\newblock \href {http://dx.doi.org/10.1145/169891.169951}
  {\path{doi:10.1145/169891.169951}}.

\bibitem{Biran2017}
O.~Biran, C.~Cotton, Explanation and justification in machine learning: A
  survey, in: Proceedings of the IJCAI 2017 Workshop on Explainable Artificial
  Intelligence (XAI), 2017, pp. 8--13.

\bibitem{Miller2019}
T.~Miller, Explanation in artificial intelligence: Insights from the social
  sciences, Artificial Intelligence 267 (2019) 1--38.
\newblock \href {http://dx.doi.org/10.1016/j.artint.2018.07.007}
  {\path{doi:10.1016/j.artint.2018.07.007}}.

\bibitem{Mittelstadt2019}
B.~D. Mittelstadt, C.~Russell, S.~Wachter, Explaining explanations in {AI}, in:
  Proceedings of the 2019 Conference on Fairness, Accountability, and
  Transparency, Association for Computing Machinery, New York, NY, USA, 2019,
  pp. 279--288.
\newblock \href {http://dx.doi.org/10.1145/3287560.3287574}
  {\path{doi:10.1145/3287560.3287574}}.

\bibitem{Burrell2016}
J.~Burrell, How the machine ‘thinks’: Understanding opacity in machine
  learning algorithms, Big Data \& Society 3~(1) (2016) 1--12.
\newblock \href {http://dx.doi.org/10.1177/2053951715622512}
  {\path{doi:10.1177/2053951715622512}}.

\bibitem{DoshiVelez2017}
F.~Doshi-Velez, B.~Kim, Towards a rigorous science of interpretable machine
  learning, CoRR abs/1702.08608.
\newblock \href {http://arxiv.org/abs/1702.08608} {\path{arXiv:1702.08608}}.

\bibitem{EUHLEGOAI}
{EU High-Level Expert Group on Artificial Intelligence},
  \href{https://ec.europa.eu/digital-single-market/en/news/ethics-guidelines-trustworthy-ai}{Ethics
  guidelines for trustworthy {AI}} (2019).
\newline\urlprefix\url{https://ec.europa.eu/digital-single-market/en/news/ethics-guidelines-trustworthy-ai}

\bibitem{Lipton2018}
Z.~C. Lipton, The mythos of model interpretability, Commun. ACM 61~(10) (2018)
  36--43.
\newblock \href {http://dx.doi.org/10.1145/3233231}
  {\path{doi:10.1145/3233231}}.

\bibitem{Adadi2018}
A.~{Adadi}, M.~{Berrada}, Peeking inside the black-box: A survey on explainable
  artificial intelligence {(XAI)}, IEEE Access 6 (2018) 52138--52160.
\newblock \href {http://dx.doi.org/10.1109/ACCESS.2018.2870052}
  {\path{doi:10.1109/ACCESS.2018.2870052}}.

\bibitem{Nunes2017}
I.~Nunes, D.~Jannach, A systematic review and taxonomy of explanations in
  decision support and recommender systems, User Modeling and User-Adapted
  Interaction 27~(3-5) (2017) 393--444.
\newblock \href {http://dx.doi.org/10.1007/s11257-017-9195-0}
  {\path{doi:10.1007/s11257-017-9195-0}}.

\bibitem{Arrieta2020}
A.~B. Arrieta, N.~Díaz-Rodríguez, J.~D. Ser, A.~Bennetot, S.~Tabik,
  A.~Barbado, S.~Garcia, S.~Gil-Lopez, D.~Molina, R.~Benjamins, R.~Chatila,
  F.~Herrera, Explainable artificial intelligence {(XAI)}: Concepts,
  taxonomies, opportunities and challenges toward responsible {AI}, Information
  Fusion 58 (2020) 82--115.
\newblock \href {http://dx.doi.org/10.1016/j.inffus.2019.12.012}
  {\path{doi:10.1016/j.inffus.2019.12.012}}.

\bibitem{Felzmann2019}
H.~Felzmann, E.~F. Villaronga, C.~Lutz, A.~Tamò-Larrieux, Transparency you can
  trust: Transparency requirements for artificial intelligence between legal
  norms and contextual concerns, Big Data \& Society 6~(1) (2019) 1--14.
\newblock \href {http://dx.doi.org/10.1177/2053951719860542}
  {\path{doi:10.1177/2053951719860542}}.

\bibitem{Gilpin2018b}
L.~H. Gilpin, C.~Testart, N.~Fruchter, J.~Adebayo, Explaining explanations to
  society, in: NIPS Workshop on Ethical, Social and Governance Issues in AI,
  Vol. abs/1901.06560, 2018, pp. 1--6.
\newblock \href {http://arxiv.org/abs/1901.06560} {\path{arXiv:1901.06560}}.

\bibitem{Hoffman2018}
R.~R. Hoffman, S.~T. Mueller, G.~Klein, J.~Litman, Metrics for explainable
  {AI}: Challenges and prospects, CoRR abs/1812.04608.
\newblock \href {http://arxiv.org/abs/1812.04608} {\path{arXiv:1812.04608}}.

\bibitem{Preece2018}
A.~D. Preece, D.~Harborne, D.~Braines, R.~Tomsett, S.~Chakraborty, Stakeholders
  in explainable {AI}, CoRR abs/1810.00184.
\newblock \href {http://arxiv.org/abs/1810.00184} {\path{arXiv:1810.00184}}.

\bibitem{Weller2019}
A.~Weller, Transparency: Motivations and challenges, in: W.~Samek, G.~Montavon,
  A.~Vedaldi, L.~K. Hansen, K.-R. M{\"u}ller (Eds.), Explainable {AI}:
  Interpreting, Explaining and Visualizing Deep Learning, Springer, 2019, pp.
  23--40.
\newblock \href {http://dx.doi.org/10.1007/978-3-030-28954-6_2}
  {\path{doi:10.1007/978-3-030-28954-6_2}}.

\bibitem{Paez2019}
A.~P{\'{a}}ez, The pragmatic turn in explainable artificial intelligence
  {(XAI)}, Minds \& Machines 29 (2019) 441--459.
\newblock \href {http://dx.doi.org/10.1007/s11023-019-09502-w}
  {\path{doi:10.1007/s11023-019-09502-w}}.

\bibitem{Cheng2019}
H.-F. Cheng, R.~Wang, Z.~Zhang, F.~O'Connell, T.~Gray, F.~M. Harper, H.~Zhu,
  Explaining decision-making algorithms through ui: Strategies to help
  non-expert stakeholders, in: Proceedings of the 2019 chi conference on human
  factors in computing systems, Association for Computing Machinery, New York,
  NY, USA, 2019, pp. 1--12.
\newblock \href {http://dx.doi.org/10.1145/3290605.3300789}
  {\path{doi:10.1145/3290605.3300789}}.

\bibitem{Abdul2018}
A.~Abdul, J.~Vermeulen, D.~Wang, B.~Y. Lim, M.~Kankanhalli, Trends and
  trajectories for explainable, accountable and intelligible systems: An {HCI}
  research agenda, in: Proceedings of the 2018 Conference on Human Factors in
  Computing Systems (CHI), Association for Computing Machinery, New York, NY,
  USA, 2018, pp. 1--18.
\newblock \href {http://dx.doi.org/10.1145/3173574.3174156}
  {\path{doi:10.1145/3173574.3174156}}.

\bibitem{MerriamStakeholder}
M.-W. Dictionary,
  \href{https://www.merriam-webster.com/dictionary/stakeholder}{Stakeholder},
  accessed: 30 July 2020 (2020).
\newline\urlprefix\url{https://www.merriam-webster.com/dictionary/stakeholder}

\bibitem{Hind2019}
M.~Hind, D.~Wei, M.~Campbell, N.~C.~F. Codella, A.~Dhurandhar,
  A.~Mojsilovi{\'c}, K.~N. Ramamurthy, K.~R. Varshney, Ted: Teaching ai to
  explain its decisions, in: Proceedings of the 2019 AAAI/ACM Conference on AI,
  Ethics, and Society, Association for Computing Machinery, New York, NY, USA,
  2019, pp. 123--129.
\newblock \href {http://dx.doi.org/10.1145/3306618.3314273}
  {\path{doi:10.1145/3306618.3314273}}.

\bibitem{Anjomshoae2019b}
S.~Anjomshoae, K.~Fr{\"a}mling, A.~Najjar, Explanations of black-box model
  predictions by contextual importance and utility, in: Explainable,
  Transparent Autonomous Agents and Multi-Agent Systems, Springer, 2019, pp.
  95--109.
\newblock \href {http://dx.doi.org/10.1007/978-3-030-30391-4_6}
  {\path{doi:10.1007/978-3-030-30391-4_6}}.

\bibitem{Atzmueller2019}
M.~Atzmueller, Towards socio-technical design of explicative systems:
  Transparent, interpretable and explainable analytics and its perspectives in
  social interaction contexts information, in: Proceedings of the 2019 Workshop
  on Affective Computing and Context Awareness in Ambient Intelligence (AfCAI),
  2019, pp. 1--8.

\bibitem{Baaj2019}
I.~Baaj, J.-P. Poli, W.~Ouerdane, Some insights towards a unified semantic
  representation of explanation for explainable artificial intelligence, in:
  Proceedings of the 2019 Workshop on Interactive Natural Language Technology
  for Explainable Artificial Intelligence (NL4XAI), Association for
  Computational Linguistics, 2019, pp. 14--19.
\newblock \href {http://dx.doi.org/10.18653/v1/W19-8404}
  {\path{doi:10.18653/v1/W19-8404}}.

\bibitem{Balog2019}
K.~Balog, F.~Radlinski, S.~Arakelyan, Transparent, scrutable and explainable
  user models for personalized recommendation, in: Proceedings of the 42nd
  International ACM SIGIR Conference on Research and Development in Information
  Retrieval, Association for Computing Machinery, New York, NY, USA, 2019, pp.
  265--274.
\newblock \href {http://dx.doi.org/10.1145/3331184.3331211}
  {\path{doi:10.1145/3331184.3331211}}.

\bibitem{Binns2018}
R.~Binns, M.~Van~Kleek, M.~Veale, U.~Lyngs, J.~Zhao, N.~Shadbolt, '{I}t's
  reducing a human being to a percentage': Perceptions of justice in
  algorithmic decisions, in: Proceedings of the 2018 Conference on Human
  Factors in Computing Systems (CHI), Association for Computing Machinery, New
  York, NY, USA, 2018, pp. 1--14.
\newblock \href {http://dx.doi.org/10.1145/3173574.3173951}
  {\path{doi:10.1145/3173574.3173951}}.

\bibitem{Chakraborti2019}
T.~Chakraborti, S.~Sreedharan, S.~Grover, S.~Kambhampati, Plan explanations as
  model reconciliation, in: 14th ACM/IEEE International Conference on
  Human-Robot Interaction (HRI), IEEE, 2019, pp. 258--266.
\newblock \href {http://dx.doi.org/10.1109/HRI.2019.8673193}
  {\path{doi:10.1109/HRI.2019.8673193}}.

\bibitem{Chen2019}
L.~Chen, D.~Yan, F.~Wang, User evaluations on sentiment-based recommendation
  explanations, ACM Transactions on Interactive Intelligent Systems (TiiS)
  9~(4) (2019) 1--38.
\newblock \href {http://dx.doi.org/10.1145/3282878}
  {\path{doi:10.1145/3282878}}.

\bibitem{Cotter2017}
K.~Cotter, J.~Cho, E.~Rader, Explaining the news feed algorithm: An analysis of
  the" news feed fyi" blog, in: Proceedings of the 2017 CHI Conference Extended
  Abstracts on Human Factors in Computing Systems (CHI EA), Association for
  Computing Machinery, New York, NY, USA, 2017, pp. 1553--1560.
\newblock \href {http://dx.doi.org/10.1145/3027063.3053114}
  {\path{doi:10.1145/3027063.3053114}}.

\bibitem{Darlington2013}
K.~Darlington, Aspects of intelligent systems explanation, Universal Journal of
  Control and Automation 1~(2) (2013) 40--51.
\newblock \href {http://dx.doi.org/10.13189/ujca.2013.010204}
  {\path{doi:10.13189/ujca.2013.010204}}.

\bibitem{Ehrlich2011}
K.~Ehrlich, S.~E. Kirk, J.~Patterson, J.~C. Rasmussen, S.~I. Ross, D.~M. Gruen,
  Taking advice from intelligent systems: The double-edged sword of
  explanations, in: Proceedings of the 16th International Conference on
  Intelligent User Interfaces (IUI), Association for Computing Machinery, New
  York, NY, USA, 2011, pp. 125--134.
\newblock \href {http://dx.doi.org/10.1145/1943403.1943424}
  {\path{doi:10.1145/1943403.1943424}}.

\bibitem{Freitas2014}
A.~A. Freitas, Comprehensible classification models: A position paper, SIGKDD
  Explorations Newsletter 15~(1) (2014) 1--10.
\newblock \href {http://dx.doi.org/10.1145/2594473.2594475}
  {\path{doi:10.1145/2594473.2594475}}.

\bibitem{Gregor1999}
S.~Gregor, I.~Benbasat, Explanations from intelligent systems: Theoretical
  foundations and implications for practice, {MIS} Quarterly 23~(4) (1999)
  497--530.
\newblock \href {http://dx.doi.org/10.2307/249487} {\path{doi:10.2307/249487}}.

\bibitem{Hois2019}
J.~Hois, D.~Theofanou-Fuelbier, A.~J. Junk, How to achieve explainability and
  transparency in human ai interaction, in: International Conference on
  Human-Computer Interaction (HCI), Springer, 2019, pp. 177--183.
\newblock \href {http://dx.doi.org/10.1007/978-3-030-23528-4_25}
  {\path{doi:10.1007/978-3-030-23528-4_25}}.

\bibitem{Kizilcec2016}
R.~F. Kizilcec, How much information? effects of transparency on trust in an
  algorithmic interface, in: Proceedings of the 2016 Conference on Human
  Factors in Computing Systems (CHI), Association for Computing Machinery, New
  York, NY, USA, 2016, pp. 2390--2395.
\newblock \href {http://dx.doi.org/10.1145/2858036.2858402}
  {\path{doi:10.1145/2858036.2858402}}.

\bibitem{Nagulendra2016}
S.~Nagulendra, J.~Vassileva, Providing awareness, explanation and control of
  personalized filtering in a social networking site, Information Systems
  Frontiers 18~(1) (2016) 145--158.
\newblock \href {http://dx.doi.org/10.1007/s10796-015-9577-y}
  {\path{doi:10.1007/s10796-015-9577-y}}.

\bibitem{Papenmeier2019}
A.~Papenmeier, G.~Englebienne, C.~Seifert, How model accuracy and explanation
  fidelity influence user trust in ai, in: Proceedings of the IJCAI 2019
  Workshop on Explainable Artificial Intelligence (XAI), 2019, pp. 94--100.

\bibitem{Pierrard2019}
R.~Pierrard, J.-P. Poli, C.~Hudelot, A new approach for explainable multiple
  organ annotation with few data, in: Proceedings of the IJCAI 2019 Workshop on
  Explainable Artificial Intelligence (XAI), 2019, pp. 101--107.

\bibitem{Putnam2019}
V.~Putnam, L.~Riegel, C.~Conati, Towards personalized xai: A case study in
  intelligent tutoring systems, in: Proceedings of the IJCAI 2019 Workshop on
  Explainable Artificial Intelligence (XAI), 2019, pp. 108--114.

\bibitem{Rader2018}
E.~Rader, K.~Cotter, J.~Cho, Explanations as mechanisms for supporting
  algorithmic transparency, in: Proceedings of the 2018 Conference on Human
  Factors in Computing Systems (CHI), Association for Computing Machinery, New
  York, NY, USA, 2018, pp. 1--13.
\newblock \href {http://dx.doi.org/10.1145/3173574.3173677}
  {\path{doi:10.1145/3173574.3173677}}.

\bibitem{Rosenfeld2019}
A.~Rosenfeld, A.~Richardson, Explainability in human--agent systems, Autonomous
  Agents and Multi-Agent Systems 33~(6) (2019) 673--705.
\newblock \href {http://dx.doi.org/10.1007/s10458-019-09408-y}
  {\path{doi:10.1007/s10458-019-09408-y}}.

\bibitem{Sato2019}
M.~Sato, K.~Nagatani, T.~Sonoda, Q.~Zhang, T.~Ohkuma, Context style explanation
  for recommender systems, Journal of Information Processing 27 (2019)
  720--729.
\newblock \href {http://dx.doi.org/10.2197/ipsjjip.27.720}
  {\path{doi:10.2197/ipsjjip.27.720}}.

\bibitem{Vig2009}
J.~Vig, S.~Sen, J.~Riedl, Tagsplanations: Explaining recommendations using
  tags, in: Proceedings of the 14th International Conference on Intelligent
  User Interfaces (IUI, Association for Computing Machinery, New York, NY, USA,
  2009, pp. 47--56.
\newblock \href {http://dx.doi.org/10.1145/1502650.1502661}
  {\path{doi:10.1145/1502650.1502661}}.

\bibitem{Watts2019}
X.~Watts, F.~L{\'e}cu{\'e}, Local score dependent model explanation for time
  dependent covariates, in: Proceedings of the IJCAI 2019 Workshop on
  Explainable Artificial Intelligence (XAI), 2019, pp. 129--135.

\bibitem{Zhou2019b}
J.~Zhou, H.~Hu, Z.~Li, K.~Yu, F.~Chen, Physiological indicators for user trust
  in machine learning with influence enhanced fact-checking, in: International
  Cross-Domain Conference for Machine Learning and Knowledge Extraction,
  Springer, 2019, pp. 94--113.
\newblock \href {http://dx.doi.org/10.1007/978-3-030-29726-8_7}
  {\path{doi:10.1007/978-3-030-29726-8_7}}.

\bibitem{Herlocker2000}
J.~L. Herlocker, J.~A. Konstan, J.~Riedl, Explaining collaborative filtering
  recommendations, in: Proceedings of the 2000 ACM Conference on Computer
  Supported Cooperative Work (CSCW), Association for Computing Machinery, New
  York, NY, USA, 2000, pp. 241--250.
\newblock \href {http://dx.doi.org/10.1145/358916.358995}
  {\path{doi:10.1145/358916.358995}}.

\bibitem{Cramer2008}
H.~Cramer, V.~Evers, S.~Ramlal, M.~Van~Someren, L.~Rutledge, N.~Stash,
  L.~Aroyo, B.~Wielinga, The effects of transparency on trust in and acceptance
  of a content-based art recommender, User Modeling and User-adapted
  interaction 18~(5) (2008) 455.
\newblock \href {http://dx.doi.org/10.1007/s11257-008-9051-3}
  {\path{doi:10.1007/s11257-008-9051-3}}.

\bibitem{Byrne2019}
R.~M. Byrne, Counterfactuals in explainable artificial intelligence (xai):
  Evidence from human reasoning, in: Proceedings of the Twenty-Eighth
  International Joint Conference on Artificial Intelligence ({IJCAI-19}), 2019,
  pp. 6276--6282.
\newblock \href {http://dx.doi.org/10.24963/ijcai.2019/876}
  {\path{doi:10.24963/ijcai.2019/876}}.

\bibitem{DeLaat2018}
P.~B. De~Laat, Algorithmic decision-making based on machine learning from big
  data: Can transparency restore accountability?, Philosophy \& Technology
  31~(4) (2018) 525--541.
\newblock \href {http://dx.doi.org/10.1007/s13347-017-0293-z}
  {\path{doi:10.1007/s13347-017-0293-z}}.

\bibitem{Floridi2018}
L.~Floridi, J.~Cowls, M.~Beltrametti, R.~Chatila, P.~Chazerand, V.~Dignum,
  C.~Luetge, R.~Madelin, U.~Pagallo, F.~Rossi, B.~Schafer, P.~Valcke,
  E.~Vayena, {AI}4people{\textemdash}{A}n ethical framework for a good {AI}
  society: Opportunities, risks, principles, and recommendations, Minds and
  Machines 28~(4) (2018) 689--707.
\newblock \href {http://dx.doi.org/10.1007/s11023-018-9482-5}
  {\path{doi:10.1007/s11023-018-9482-5}}.

\bibitem{Lepri2018}
B.~Lepri, N.~Oliver, E.~Letouz{\'e}, A.~Pentland, P.~Vinck, Fair, transparent,
  and accountable algorithmic decision-making processes, Philosophy \&
  Technology 31~(4) (2018) 611--627.
\newblock \href {http://dx.doi.org/10.1007/s13347-017-0279-x}
  {\path{doi:10.1007/s13347-017-0279-x}}.

\bibitem{Mathews2019}
S.~M. Mathews, Explainable artificial intelligence applications in nlp,
  biomedical, and malware classification: A literature review, in: Intelligent
  Computing -- Proceedings of the Computing Conference, Springer, 2019, pp.
  1269--1292.
\newblock \href {http://dx.doi.org/10.1007/978-3-030-22868-2_90}
  {\path{doi:10.1007/978-3-030-22868-2_90}}.

\bibitem{Mittelstadt2016}
B.~D. Mittelstadt, P.~Allo, M.~Taddeo, S.~Wachter, L.~Floridi, The ethics of
  algorithms: Mapping the debate, Big Data \& Society 3~(2) (2016) 1--21.
\newblock \href {http://dx.doi.org/10.1177/2053951716679679}
  {\path{doi:10.1177/2053951716679679}}.

\bibitem{Pieters2011}
W.~Pieters, Explanation and trust: what to tell the user in security and ai?,
  Ethics and Information Technology 13~(1) (2011) 53--64.
\newblock \href {http://dx.doi.org/10.1007/s10676-010-9253-3}
  {\path{doi:10.1007/s10676-010-9253-3}}.

\bibitem{Ras2018}
G.~Ras, M.~van Gerven, P.~Haselager, Explanation methods in deep learning:
  Users, values, concerns and challenges, in: Explainable and Interpretable
  Models in Computer Vision and Machine Learning, Springer, 2018, pp. 19--36.
\newblock \href {http://dx.doi.org/10.1007/978-3-319-98131-4_2}
  {\path{doi:10.1007/978-3-319-98131-4_2}}.

\bibitem{Riedl2019}
M.~O. Riedl, Human-centered artificial intelligence and machine learning, Human
  Behavior and Emerging Technologies 1~(1) (2019) 33--36.
\newblock \href {http://dx.doi.org/10.1002/hbe2.117}
  {\path{doi:10.1002/hbe2.117}}.

\bibitem{Robbins2019}
S.~Robbins, A misdirected principle with a catch: Explicability for ai, Minds
  and Machines 29~(4) (2019) 495--514.
\newblock \href {http://dx.doi.org/10.1007/s11023-019-09509-3}
  {\path{doi:10.1007/s11023-019-09509-3}}.

\bibitem{Sheh2017}
R.~Sheh, Different xai for different hri, in: AAAI Fall Symposium, {AAAI}
  Press, 2017, pp. 114--117.

\bibitem{Sheh2018}
R.~Sheh, I.~Monteath, Defining explainable ai for requirements analysis,
  KI-K{\"u}nstliche Intelligenz 32~(4) (2018) 261--266.
\newblock \href {http://dx.doi.org/10.1007/s13218-018-0559-3}
  {\path{doi:10.1007/s13218-018-0559-3}}.

\bibitem{Sokol2020}
K.~Sokol, P.~A. Flach, Explainability fact sheets: A framework for systematic
  assessment of explainable approaches, in: Proceedings of the 2020 Conference
  on Fairness, Accountability, and Transparency, Association for Computing
  Machinery, New York, NY, USA, 2020, pp. 56--67.
\newblock \href {http://dx.doi.org/10.1145/3351095.3372870}
  {\path{doi:10.1145/3351095.3372870}}.

\bibitem{Sokol2020b}
K.~Sokol, P.~Flach, One explanation does not fit all, KI-K{\"u}nstliche
  Intelligenz 34~(2) (2020) 235--250.
\newblock \href {http://dx.doi.org/10.1007/s13218-020-00637-y}
  {\path{doi:10.1007/s13218-020-00637-y}}.

\bibitem{Sridharan2019}
M.~Sridharan, B.~Meadows, Towards a theory of explanations for human--robot
  collaboration, KI-K{\"u}nstliche Intelligenz 33~(4) (2019) 331--342.
\newblock \href {http://dx.doi.org/10.1007/s13218-019-00616-y}
  {\path{doi:10.1007/s13218-019-00616-y}}.

\bibitem{Vellido2019}
A.~Vellido, The importance of interpretability and visualization in machine
  learning for applications in medicine and health care, Neural Computing and
  Applications\href {http://dx.doi.org/10.1007/s00521-019-04051-w}
  {\path{doi:10.1007/s00521-019-04051-w}}.

\bibitem{Wang2019}
D.~Wang, Q.~Yang, A.~Abdul, B.~Y. Lim, Designing theory-driven user-centric
  explainable ai, in: Proceedings of the 2019 Conference on Human Factors in
  Computing Systems (CHI), Association for Computing Machinery, New York, NY,
  USA, 2019, pp. 1--15.
\newblock \href {http://dx.doi.org/10.1145/3290605.3300831}
  {\path{doi:10.1145/3290605.3300831}}.

\bibitem{Lee2019}
M.~K. Lee, A.~Jain, H.~J. Cha, S.~Ojha, D.~Kusbit, Procedural justice in
  algorithmic fairness, Proceedings of the 2019 {ACM} on Human-Computer
  Interaction 3 (2019) 1--26.
\newblock \href {http://dx.doi.org/10.1145/3359284}
  {\path{doi:10.1145/3359284}}.

\bibitem{Doran2018}
D.~Doran, S.~Schulz, T.~R. Besold, What does explainable ai really mean? a new
  conceptualization of perspectives, in: CEUR Workshop Proceedings, Vol. 2071,
  CEUR, 2018, pp. 1--8.
\newblock \href {http://arxiv.org/abs/1710.00794} {\path{arXiv:1710.00794}}.

\bibitem{Krishnan2019}
M.~Krishnan, Against interpretability: A critical examination of the
  interpretability problem in machine learning, Philosophy \& Technology (2019)
  1--16\href {http://dx.doi.org/10.1007/s13347-019-00372-9}
  {\path{doi:10.1007/s13347-019-00372-9}}.

\bibitem{Peddoju2019}
S.~K. Peddoju, M.~Saravanan, S.~Suresh, Explainable classification using
  clustering in deep learning models, in: Proceedings of the IJCAI 2019
  Workshop on Explainable Artificial Intelligence (XAI), 2019, pp. 115--121.

\bibitem{Rajani2017}
N.~F. Rajani, R.~J. Mooney, Using explanations to improve ensembling of visual
  question answering systems, in: Proceedings of the IJCAI 2017 Workshop on
  Explainable Artificial Intelligence (XAI), 2017, pp. 43--47.

\bibitem{Zhou2019}
J.~Zhou, F.~Chen, Towards trustworthy human-ai teaming under uncertainty, in:
  Proceedings of the IJCAI 2019 Workshop on Explainable Artificial Intelligence
  (XAI), 2019, pp. 143--147.

\bibitem{Anjomshoae2019}
S.~Anjomshoae, A.~Najjar, D.~Calvaresi, K.~Fr\"{a}mling, Explainable agents and
  robots: Results from a systematic literature review, in: Proceedings of the
  18th International Conference on Autonomous Agents and MultiAgent Systems
  (AAMAS), International Foundation for Autonomous Agents and Multiagent
  Systems, Richland, SC, USA, 2019, p. 1078–1088.
\newblock \href {http://dx.doi.org/10.5555/3306127.3331806}
  {\path{doi:10.5555/3306127.3331806}}.

\bibitem{Fox2017}
M.~Fox, D.~Long, D.~Magazzeni, Explainable planning, in: Proceedings of the
  IJCAI 2017 Workshop on Explainable Artificial Intelligence (XAI), 2017, pp.
  24--30.

\bibitem{Jasanoff2017}
S.~Jasanoff, Virtual, visible, and actionable: Data assemblages and the
  sightlines of justice, Big Data \& Society 4~(2) (2017) 1--15.
\newblock \href {http://dx.doi.org/10.1177/2053951717724477}
  {\path{doi:10.1177/2053951717724477}}.

\bibitem{Friedrich2011}
G.~Friedrich, M.~Zanker, A taxonomy for generating explanations in recommender
  systems, AI Magazine 32~(3) (2011) 90--98.
\newblock \href {http://dx.doi.org/10.1609/aimag.v32i3.2365}
  {\path{doi:10.1609/aimag.v32i3.2365}}.

\bibitem{Holzinger2019}
A.~Holzinger, G.~Langs, H.~Denk, K.~Zatloukal, H.~M{\"u}ller, Causability and
  explainability of artificial intelligence in medicine, Wiley
  Interdisciplinary Reviews: Data Mining and Knowledge Discovery 9~(4) (2019)
  1--13.
\newblock \href {http://dx.doi.org/10.1002/widm.1312}
  {\path{doi:10.1002/widm.1312}}.

\bibitem{Sevastjanova2018}
R.~Sevastjanova, F.~Beck, B.~Ell, C.~Turkay, R.~Henkin, M.~Butt, D.~A. Keim,
  M.~El-Assady, Going beyond visualization: Verbalization as complementary
  medium to explain machine learning models, in: VIS Workshop on Visualization
  for AI Explainability (VISxAI), 2018, pp. 1--6.

\bibitem{Sormo2005}
F.~S{\o}rmo, J.~Cassens, A.~Aamodt, Explanation in case-based
  reasoning--perspectives and goals, Artificial Intelligence Review 24~(2)
  (2005) 109--143.
\newblock \href {http://dx.doi.org/10.1007/s10462-005-4607-7}
  {\path{doi:10.1007/s10462-005-4607-7}}.

\bibitem{Zerilli2019}
J.~Zerilli, A.~Knott, J.~Maclaurin, C.~Gavaghan, Transparency in algorithmic
  and human decision-making: is there a double standard?, Philosophy \&
  Technology 32~(4) (2019) 661--683.
\newblock \href {http://dx.doi.org/10.1007/s13347-018-0330-6}
  {\path{doi:10.1007/s13347-018-0330-6}}.

\bibitem{Lucic2019}
A.~Lucic, H.~Haned, M.~de~Rijke, Contrastive explanations for large errors in
  retail forecasting predictions through monte carlo simulations, in:
  Proceedings of the IJCAI 2019 Workshop on Explainable Artificial Intelligence
  (XAI), 2019, pp. 66--72.

\bibitem{Dam2018}
H.~K. Dam, T.~Tran, A.~Ghose, Explainable software analytics, in: Proceedings
  of the 40th International Conference on Software Engineering: New Ideas and
  Emerging Results (ICSE-NIER), Association for Computing Machinery, New York,
  NY, USA, 2018, pp. 53--56.
\newblock \href {http://dx.doi.org/10.1145/3183399.3183424}
  {\path{doi:10.1145/3183399.3183424}}.

\bibitem{DeWinter2010}
J.~De~Winter, Explanations in software engineering: The pragmatic point of
  view, Minds and Machines 20~(2) (2010) 277--289.
\newblock \href {http://dx.doi.org/10.1007/s11023-010-9190-2}
  {\path{doi:10.1007/s11023-010-9190-2}}.

\bibitem{Juozapaitis2019}
Z.~Juozapaitis, A.~Koul, A.~Fern, M.~Erwig, F.~Doshi-Velez, Explainable
  reinforcement learning via reward decomposition, in: Proceedings of the IJCAI
  2019 Workshop on Explainable Artificial Intelligence (XAI), 2019, pp. 47--53.

\bibitem{Michael2019}
L.~Michael, Machine coaching, in: Proceedings of the IJCAI 2019 Workshop on
  Explainable Artificial Intelligence (XAI), 2019, pp. 80--86.

\bibitem{Sokol2018}
K.~Sokol, P.~A. Flach, Conversational explanations of machine learning
  predictions through class-contrastive counterfactual statements, in:
  Proceedings of the Twenty-Seventh International Joint Conference on
  Artificial Intelligence ({IJCAI-18}), 2018, pp. 5785--5786.
\newblock \href {http://dx.doi.org/10.24963/ijcai.2018/836}
  {\path{doi:10.24963/ijcai.2018/836}}.

\bibitem{Wicaksono2017}
H.~Wicaksono, C.~Sammut, R.~Sheh, Towards explainable tool creation by a robot,
  in: Proceedings of the IJCAI 2017 Workshop on Explainable Artificial
  Intelligence (XAI), 2017, pp. 63--67.

\bibitem{Eiter2019}
T.~Eiter, Z.~G. Saribatur, P.~Sch{\"u}ller, Abstraction for zooming-in to
  unsolvability reasons of grid-cell problems, in: Proceedings of the IJCAI
  2019 Workshop on Explainable Artificial Intelligence (XAI), 2019, pp. 7--13.

\bibitem{Kulesza2011}
T.~Kulesza, S.~Stumpf, W.-K. Wong, M.~M. Burnett, S.~Perona, A.~Ko, I.~Oberst,
  Why-oriented end-user debugging of naive bayes text classification, ACM
  Transactions on Interactive Intelligent Systems (TiiS) 1~(1) (2011) 1--31.
\newblock \href {http://dx.doi.org/10.1145/2030365.2030367}
  {\path{doi:10.1145/2030365.2030367}}.

\bibitem{Hoffman2018b}
R.~R. Hoffman, G.~Klein, S.~T. Mueller, Explaining explanation for
  “explainable ai”, Proceedings of the Human Factors and Ergonomics Society
  Annual Meeting 62~(1) (2018) 197--201.
\newblock \href {http://dx.doi.org/10.1177/1541931218621047}
  {\path{doi:10.1177/1541931218621047}}.

\bibitem{Nothdurft2013}
F.~Nothdurft, T.~Heinroth, W.~Minker, The impact of explanation dialogues on
  human-computer trust, in: International Conference on Human-Computer
  Interaction (HCI), Springer, 2013, pp. 59--67.
\newblock \href {http://dx.doi.org/10.1007/978-3-642-39265-8_7}
  {\path{doi:10.1007/978-3-642-39265-8_7}}.

\bibitem{Brinton2017}
C.~Brinton, A framework for explanation of machine learning decisions, in:
  Proceedings of the IJCAI 2017 Workshop on Explainable Artificial Intelligence
  (XAI), 2017, pp. 14--18.

\bibitem{Tintarev2007}
N.~Tintarev, Explanations of recommendations, in: Proceedings of the 2007 ACM
  Conference on Recommender Systems, Association for Computing Machinery, New
  York, NY, USA, 2007, pp. 203--206.
\newblock \href {http://dx.doi.org/10.1145/1297231.1297275}
  {\path{doi:10.1145/1297231.1297275}}.

\bibitem{Weber2019}
R.~O. Weber, H.~Hong, P.~Goel, Explaining citation recommendations: Abstracts
  or full texts?, in: Proceedings of the IJCAI 2019 Workshop on Explainable
  Artificial Intelligence (XAI), 2019, pp. 136--142.

\bibitem{Gilpin2018}
L.~H. Gilpin, D.~Bau, B.~Z. Yuan, A.~Bajwa, M.~Specter, L.~Kagal, Explaining
  explanations: An overview of interpretability of machine learning, in: IEEE
  5th International Conference on Data Science and Advanced Analytics {DSAA},
  2018, pp. 80--89.
\newblock \href {http://dx.doi.org/10.1109/DSAA.2018.00018}
  {\path{doi:10.1109/DSAA.2018.00018}}.

\bibitem{Ho2020}
J.~C.-T. Ho, How biased is the sample? reverse engineering the ranking
  algorithm of facebook’s graph application programming interface, Big Data
  \& Society 7~(1) (2020) 1--15.
\newblock \href {http://dx.doi.org/10.1177/2053951720905874}
  {\path{doi:10.1177/2053951720905874}}.

\bibitem{Hohman2019}
F.~Hohman, A.~Head, R.~Caruana, R.~DeLine, S.~M. Drucker, Gamut: A design probe
  to understand how data scientists understand machine learning models, in:
  Proceedings of the 2019 Conference on Human Factors in Computing Systems
  (CHI), Association for Computing Machinery, New York, NY, USA, 2019, pp.
  1--13.
\newblock \href {http://dx.doi.org/10.1145/3290605.3300809}
  {\path{doi:10.1145/3290605.3300809}}.

\bibitem{Veale2017}
M.~Veale, R.~Binns, Fairer machine learning in the real world: Mitigating
  discrimination without collecting sensitive data, Big Data \& Society 4~(2)
  (2017) 1--17.
\newblock \href {http://dx.doi.org/10.1177/2053951717743530}
  {\path{doi:10.1177/2053951717743530}}.

\bibitem{Zednik2019}
C.~Zednik, Solving the black box problem: A normative framework for explainable
  artificial intelligence, Philosophy \& Technology (2019) 1--24\href
  {http://dx.doi.org/10.1007/s13347-019-00382-7}
  {\path{doi:10.1007/s13347-019-00382-7}}.

\bibitem{Goodman2017}
B.~Goodman, S.~Flaxman, European union regulations on algorithmic
  decision-making and a "right to explanation", AI magazine 38~(3) (2017)
  50--57.
\newblock \href {http://dx.doi.org/10.1609/aimag.v38i3.2741}
  {\path{doi:10.1609/aimag.v38i3.2741}}.

\bibitem{Sklar2018}
E.~I. Sklar, M.~Q. Azhar, Explanation through argumentation, in: Proceedings of
  the 6th International Conference on Human-Agent Interaction (HAI),
  Association for Computing Machinery, New York, NY, USA, 2018, pp. 277--285.
\newblock \href {http://dx.doi.org/10.1145/3284432.3284470}
  {\path{doi:10.1145/3284432.3284470}}.

\bibitem{Lage2019}
I.~Lage, D.~Lifschitz, F.~Doshi-Velez, O.~Amir, Exploring computational user
  models for agent policy summarization, in: Proceedings of the IJCAI 2019
  Workshop on Explainable Artificial Intelligence (XAI), 2019, pp. 59--65.

\bibitem{Dahl2018}
E.~S. Dahl, Appraising black-boxed technology: the positive prospects,
  Philosophy \& Technology 31~(4) (2018) 571--591.
\newblock \href {http://dx.doi.org/10.1007/s13347-017-0275-1}
  {\path{doi:10.1007/s13347-017-0275-1}}.

\bibitem{Ghosh2019}
B.~Ghosh, D.~Malioutov, K.~S. Meel, Interpretable classification rules in
  relaxed logical form, in: Proceedings of the IJCAI 2019 Workshop on
  Explainable Artificial Intelligence (XAI), 2019, pp. 14--20.

\bibitem{Stuart2019}
M.~T. Stuart, N.~J. Nersessian, Peeking inside the black box: a new kind of
  scientific visualization, Minds and Machines 29~(1) (2019) 87--107.
\newblock \href {http://dx.doi.org/10.1007/s11023-018-9484-3}
  {\path{doi:10.1007/s11023-018-9484-3}}.

\bibitem{Clos2017}
J.~Clos, N.~Wiratunga, S.~Massie, Towards explainable text classification by
  jointly learning lexicon and modifier terms, in: Proceedings of the IJCAI
  2017 Workshop on Explainable Artificial Intelligence (XAI), 2017, pp. 19--23.

\bibitem{Zhu2018}
J.~Zhu, A.~Liapis, S.~Risi, R.~Bidarra, G.~M. Youngblood, Explainable ai for
  designers: A human-centered perspective on mixed-initiative co-creation, in:
  IEEE Conference on Computational Intelligence and Games (CIG), IEEE, 2018,
  pp. 1--8.
\newblock \href {http://dx.doi.org/10.1109/CIG.2018.8490433}
  {\path{doi:10.1109/CIG.2018.8490433}}.

\bibitem{Clinciu2019}
M.-A. Clinciu, H.~Hastie, A survey of explainable ai terminology, in:
  Proceedings of the 1st Workshop on Interactive Natural Language Technology
  for Explainable Artificial Intelligence (NL4XAI 2019), Association for
  Computational Linguistics, 2019, pp. 8--13.
\newblock \href {http://dx.doi.org/10.18653/v1/W19-8403}
  {\path{doi:10.18653/v1/W19-8403}}.

\bibitem{Henin2019}
C.~Henin, D.~Le~M{\'e}tayer, Towards a generic framework for black-box
  explanation methods, in: Proceedings of the IJCAI 2019 Workshop on
  Explainable Artificial Intelligence (XAI), 2019, pp. 28--34.

\bibitem{Madumal2019b}
P.~Madumal, T.~Miller, L.~Sonenberg, F.~Vetere, A grounded interaction protocol
  for explainable artificial intelligence, in: Proceedings of the 18th
  International Conference on Autonomous Agents and MultiAgent Systems (AAMAS),
  International Foundation for Autonomous Agents and Multiagent Systems,
  Richland, SC, USA, 2019, pp. 1033--1041.
\newblock \href {http://dx.doi.org/10.5555/3306127.3331801}
  {\path{doi:10.5555/3306127.3331801}}.

\bibitem{Olson2019}
M.~L. Olson, L.~Neal, F.~Li, W.-K. Wong, Counterfactual states for atari agents
  via generative deep learning, in: Proceedings of the IJCAI 2019 Workshop on
  Explainable Artificial Intelligence (XAI), 2019, pp. 87--93.

\bibitem{Zeng2018}
Z.~Zeng, C.~Miao, C.~Leung, J.~J. Chin,
  \href{https://www.aaai.org/ocs/index.php/AAAI/AAAI18/paper/view/16762}{Building
  more explainable artificial intelligence with argumentation}, in: Proceedings
  of the Thirty-Second {AAAI} Conference on Artificial Intelligence (AAAI-18),
  {AAAI} Press, 2018, pp. 8044--8046.
\newline\urlprefix\url{https://www.aaai.org/ocs/index.php/AAAI/AAAI18/paper/view/16762}

\bibitem{Madumal2019a}
P.~Madumal, T.~Miller, L.~Sonenberg, F.~Vetere, Explainable reinforcement
  learning through a causal lens, in: Proceedings of the IJCAI 2019 Workshop on
  Explainable Artificial Intelligence (XAI), 2019, pp. 73--79.

\bibitem{Ribeiro2016}
M.~T. Ribeiro, S.~Singh, C.~Guestrin, "{W}hy should {I} trust you?": Explaining
  the predictions of any classifier, in: Proceedings of the 22nd {ACM} {SIGKDD}
  International Conference on Knowledge Discovery and Data Mining, Association
  for Computing Machinery, New York, NY, USA, 2016, pp. 1135--1144.
\newblock \href {http://dx.doi.org/10.1145/2939672.2939778}
  {\path{doi:10.1145/2939672.2939778}}.

\bibitem{Endsley2017}
M.~R. Endsley, From here to autonomy, Human Factors: The Journal of the Human
  Factors and Ergonomics Society 59~(1) (2017) 5--27.
\newblock \href {http://dx.doi.org/10.1177/0018720816681350}
  {\path{doi:10.1177/0018720816681350}}.

\bibitem{Lee2004}
J.~D. Lee, K.~A. See, Trust in automation: Designing for appropriate reliance,
  Human Factors 46~(1) (2004) 50--80.
\newblock \href {http://dx.doi.org/10.1518/hfes.46.1.50_30392}
  {\path{doi:10.1518/hfes.46.1.50_30392}}.

\bibitem{Parasuraman1997}
R.~Parasuraman, V.~Riley, Humans and automation: Use, misuse, disuse, abuse,
  Human Factors: The Journal of the Human Factors and Ergonomics Society 39~(2)
  (1997) 230--253.
\newblock \href {http://dx.doi.org/10.1518/001872097778543886}
  {\path{doi:10.1518/001872097778543886}}.

\bibitem{Hoff2014}
K.~A. Hoff, M.~Bashir, Trust in automation, Human Factors 57~(3) (2014)
  407--434.
\newblock \href {http://dx.doi.org/10.1177/0018720814547570}
  {\path{doi:10.1177/0018720814547570}}.

\bibitem{Parasuraman2010}
R.~Parasuraman, D.~H. Manzey, Complacency and bias in human use of automation:
  An attentional integration, Human Factors 52~(3) (2010) 381--410.
\newblock \href {http://dx.doi.org/10.1177/0018720810376055}
  {\path{doi:10.1177/0018720810376055}}.

\bibitem{Kunze2019}
A.~Kunze, S.~J. Summerskill, R.~Marshall, A.~J. Filtness, Automation
  transparency: Implications of uncertainty communication for human-automation
  interaction and interfaces, Ergonomics 62~(3) (2019) 345--360.
\newblock \href {http://dx.doi.org/10.1080/00140139.2018.1547842}
  {\path{doi:10.1080/00140139.2018.1547842}}.

\bibitem{Samek2017}
W.~Samek, T.~Wiegand, K.-R. M{\"u}ller, Explainable artificial intelligence:
  Understanding, visualizing and interpreting deep learning models, CoRR
  abs/1708.08296.
\newblock \href {http://arxiv.org/abs/1708.08296} {\path{arXiv:1708.08296}}.

\bibitem{Montavon2018}
G.~Montavon, W.~Samek, K.-R. M{\"u}ller, Methods for interpreting and
  understanding deep neural networks, Digital Signal Processing 73 (2018)
  1--15.
\newblock \href {http://dx.doi.org/10.1016/j.dsp.2017.10.011}
  {\path{doi:10.1016/j.dsp.2017.10.011}}.

\bibitem{Becker2018}
S.~Becker, M.~Ackermann, S.~Lapuschkin, K.~M{\"{u}}ller, W.~Samek, Interpreting
  and explaining deep neural networks for classification of audio signals, CoRR
  abs/1807.03418.
\newblock \href {http://arxiv.org/abs/1807.03418} {\path{arXiv:1807.03418}}.

\bibitem{Lapuschkin2016}
S.~Lapuschkin, A.~Binder, G.~Montavon, K.-R. Muller, W.~Samek, Analyzing
  classifiers: Fisher vectors and deep neural networks, in: Proceedings of the
  IEEE Conference on Computer Vision and Pattern Recognition, 2016, pp.
  2912--2920.
\newblock \href {http://dx.doi.org/10.1109/CVPR.2016.318}
  {\path{doi:10.1109/CVPR.2016.318}}.

\bibitem{Caruana2015}
R.~Caruana, Y.~Lou, J.~Gehrke, P.~Koch, M.~Sturm, N.~Elhadad, Intelligible
  models for healthcare: Predicting pneumonia risk and hospital 30-day
  readmission, in: Proceedings of the 21th ACM SIGKDD International Conference
  on Knowledge Discovery and Data Mining, Association for Computing Machinery,
  New York, NY, USA, 2015, pp. 1721--1730.
\newblock \href {http://dx.doi.org/10.1145/2783258.2788613}
  {\path{doi:10.1145/2783258.2788613}}.

\bibitem{Baum2018}
K.~Baum, H.~Hermanns, T.~Speith,
  \href{http://isaim2018.cs.virginia.edu/papers/ISAIM2018_Ethics_Baum_etal.pdf}{From
  machine ethics to machine explainability and back}, in: International
  Symposium on Artificial Intelligence and Mathematics ({ISAIM}), 2018, pp.
  1--8.
\newline\urlprefix\url{http://isaim2018.cs.virginia.edu/papers/ISAIM2018_Ethics_Baum_etal.pdf}

\bibitem{Luetge2017}
C.~Luetge, The german ethics code for automated and connected driving,
  Philosophy \& Technology 30 (2017) 547--558.
\newblock \href {http://dx.doi.org/10.1007/s13347-017-0284-0}
  {\path{doi:10.1007/s13347-017-0284-0}}.

\bibitem{Purkiss2006}
S.~L.~S. Purkiss, P.~L. Perrewé, T.~L. Gillespie, B.~T. Mayes, G.~R. Ferris,
  Implicit sources of bias in employment interview judgments and decisions,
  Organizational Behavior and Human Decision Processes 101~(2) (2006) 152--167.
\newblock \href {http://dx.doi.org/10.1016/j.obhdp.2006.06.005}
  {\path{doi:10.1016/j.obhdp.2006.06.005}}.

\bibitem{Caliskan2017}
A.~Caliskan, J.~J. Bryson, A.~Narayanan, Semantics derived automatically from
  language corpora contain human-like biases, Science 356 (2017) 183--186.
\newblock \href {http://dx.doi.org/10.1126/science.aal4230}
  {\path{doi:10.1126/science.aal4230}}.

\bibitem{Guidotti2019}
R.~Guidotti, A.~Monreale, S.~Ruggieri, F.~Turini, F.~Giannotti, D.~Pedreschi, A
  survey of methods for explaining black box models, {ACM} Comput. Surv. 51~(5)
  (2019) 1--42.
\newblock \href {http://dx.doi.org/10.1145/3236009}
  {\path{doi:10.1145/3236009}}.

\bibitem{Venkatesh2003}
V.~Venkatesh, M.~G. Morris, G.~B. Davis, F.~D. Davis, User acceptance of
  information technology: Toward a unified view, Management Information Systems
  Quarterly 27~(3) (2003) 425--478.
\newblock \href {http://dx.doi.org/10.2307/30036540}
  {\path{doi:10.2307/30036540}}.

\bibitem{McLeod2015}
C.~McLeod, Trust, in: E.~N. Zalta (Ed.), The Stanford Encyclopedia of
  Philosophy, fall 2015 Edition, Metaphysics Research Lab, Stanford University,
  2015, pp. 1--43.

\bibitem{Raji2020}
I.~D. Raji, A.~Smart, R.~N. White, M.~Mitchell, T.~Gebru, B.~Hutchinson,
  J.~Smith-Loud, D.~Theron, P.~Barnes, Closing the ai accountability gap:
  Defining an end-to-end framework for internal algorithmic auditing, in:
  Proceedings of the 2020 Conference on Fairness, Accountability, and
  Transparency (FAT*), Association for Computing Machinery, New York, NY, USA,
  2020, pp. 33--44.
\newblock \href {http://dx.doi.org/10.1145/3351095.3372873}
  {\path{doi:10.1145/3351095.3372873}}.

\bibitem{Matthias2004}
A.~Matthias, The responsibility gap: Ascribing responsibility for the actions
  of learning automata, Ethics and information technology 6~(3) (2004)
  175--183.
\newblock \href {http://dx.doi.org/10.1007/s10676-004-3422-1}
  {\path{doi:10.1007/s10676-004-3422-1}}.

\bibitem{Deci2017}
E.~L. Deci, A.~H. Olafsen, R.~M. Ryan, Self-determination theory in work
  organizations: The state of a science, Annual Review of Organizational
  Psychology and Organizational Behavior 4~(1) (2017) 19--43.
\newblock \href {http://dx.doi.org/10.1146/annurev-orgpsych-032516-113108}
  {\path{doi:10.1146/annurev-orgpsych-032516-113108}}.

\bibitem{Longoni2019}
C.~Longoni, A.~Bonezzi, C.~K. Morewedge, Resistance to medical artificial
  intelligence, Journal of Consumer Research 46~(4) (2019) 629--650.
\newblock \href {http://dx.doi.org/10.1093/jcr/ucz013}
  {\path{doi:10.1093/jcr/ucz013}}.

\bibitem{Keil2006}
F.~C. Keil, Explanation and understanding, Annual Review of Psychology 57~(1)
  (2006) 227--254.
\newblock \href {http://dx.doi.org/10.1146/annurev.psych.57.102904.190100}
  {\path{doi:10.1146/annurev.psych.57.102904.190100}}.

\bibitem{Bonnefon2016}
J.-F. Bonnefon, A.~Shariff, I.~Rahwan, The social dilemma of autonomous
  vehicles, Science 352 (2016) 1573--1576.
\newblock \href {http://dx.doi.org/10.1126/science.aaf2654}
  {\path{doi:10.1126/science.aaf2654}}.

\bibitem{Buchanan1984}
B.~G. Buchanan, E.~H. Shortliffe, Rule-based expert systems: The MYCIN
  experiments of the Stanford Heuristic Programming Project, Addison-Wesley,
  1984.

\bibitem{Dhaliwal1996}
J.~S. Dhaliwal, I.~Benbasat, The use and effects of knowledge-based system
  explanations: Theoretical foundations and a framework for empirical
  evaluation, Information Systems Research 7~(3) (1996) 342--362.
\newblock \href {http://dx.doi.org/10.1287/isre.7.3.342}
  {\path{doi:10.1287/isre.7.3.342}}.

\bibitem{Koehl2019}
M.~A. Köhl, K.~Baum, M.~Langer, D.~Oster, T.~Speith, D.~Bohlender,
  Explainability as a non-functional requirement, in: IEEE 27th International
  Requirements Engineering Conference (RE), 2019, pp. 363--368.
\newblock \href {http://dx.doi.org/10.1109/RE.2019.00046}
  {\path{doi:10.1109/RE.2019.00046}}.

\bibitem{DeRegt2017}
H.~W. De~Regt, Understanding scientific understanding, Oxford University Press,
  2017.
\newblock \href {http://dx.doi.org/10.1093/oso/9780190652913.001.0001}
  {\path{doi:10.1093/oso/9780190652913.001.0001}}.

\bibitem{Baumberger2017}
C.~Baumberger, C.~Beisbart, G.~Brun, What is understanding? {A}n overview of
  recent debates in epistemology and philosophy of science, in: S.~G.~C.
  Baumberger, S.~Ammon (Eds.), Explaining Understanding: New Perspectives from
  Epistemolgy and Philosophy of Science, Routledge, 2017, pp. 1--34.

\bibitem{Malfatti2019}
F.~I. Malfatti, On understanding and testimony, Erkenntnis (2019) 1--21\href
  {http://dx.doi.org/10.1007/s10670-019-00157-8}
  {\path{doi:10.1007/s10670-019-00157-8}}.

\bibitem{Baumberger2014}
C.~Baumberger, Types of understanding: Their nature and their relation to
  knowledge, Conceptus 40~(98) (2014) 67--88.
\newblock \href {http://dx.doi.org/10.1515/cpt-2014-0002}
  {\path{doi:10.1515/cpt-2014-0002}}.

\bibitem{Lambert1991}
K.~Lambert, On whether an answer to a why-question is an explanation if and
  only if it yields scientific understanding, in: G.~G. Brittan (Ed.),
  Causality, method, and modality, Vol.~48, Springer, Dordrecht, Netherlands,
  1991, pp. 125--142.
\newblock \href {http://dx.doi.org/10.1007/978-94-011-3348-7_8}
  {\path{doi:10.1007/978-94-011-3348-7_8}}.

\bibitem{Lombrozo2006}
T.~Lombrozo, S.~Carey, Functional explanation and the function of explanation,
  Cognition 99~(2) (2006) 167--204.
\newblock \href {http://dx.doi.org/10.1016/j.cognition.2004.12.009}
  {\path{doi:10.1016/j.cognition.2004.12.009}}.

\bibitem{Chi1994}
M.~T. Chi, N.~De~Leeuw, M.-H. Chiu, C.~Lavancher, Eliciting self-explanations
  improves understanding, Cognitive Science 18~(3) (1994) 439--477.
\newblock \href {http://dx.doi.org/10.1207/s15516709cog1803\_3}
  {\path{doi:10.1207/s15516709cog1803\_3}}.

\bibitem{Mayer1992}
R.~E. Mayer, Cognition and instruction: Their historic meeting within
  educational psychology, Journal of educational Psychology 84~(4) (1992)
  405--412.
\newblock \href {http://dx.doi.org/10.1037/0022-0663.84.4.405}
  {\path{doi:10.1037/0022-0663.84.4.405}}.

\bibitem{Mueller2019}
S.~T. Mueller, R.~R. Hoffman, W.~Clancey, A.~Emrey, G.~Klein, Explanation in
  human-ai systems: A literature meta-review, synopsis of key ideas and
  publications, and bibliography for explainable ai, CoRR abs/1902.01876.
\newblock \href {http://arxiv.org/abs/1902.01876} {\path{arXiv:1902.01876}}.

\bibitem{Kelp2015}
C.~Kelp, Understanding phenomena, Synthese 192~(12) (2015) 3799--3816.
\newblock \href {http://dx.doi.org/10.1007/s11229-014-0616-x}
  {\path{doi:10.1007/s11229-014-0616-x}}.

\bibitem{Feltovich2001}
P.~J. Feltovich, R.~L. Coulson, R.~J. Spiro, Learners' (mis)understanding of
  important and difficult concepts: A challenge to smart machines in education,
  in: Smart Machines in Education: The Coming Revolution in Educational
  Technology, MIT Press, Cambridge, MA, USA, 2001, pp. 349--375.

\bibitem{Rouse1986}
W.~B. Rouse, N.~M. Morris, On looking into the black box: Prospects and limits
  in the search for mental models, Psychological bulletin 100~(3) (1986)
  349--363.
\newblock \href {http://dx.doi.org/10.1037/0033-2909.100.3.349}
  {\path{doi:10.1037/0033-2909.100.3.349}}.

\bibitem{Rozenblit2002}
L.~Rozenblit, F.~Keil, The misunderstood limits of folk science: An illusion of
  explanatory depth, Cognitive Science 26~(5) (2002) 521--562.
\newblock \href {http://dx.doi.org/10.1207/s15516709cog2605_1}
  {\path{doi:10.1207/s15516709cog2605_1}}.

\bibitem{Kuhn2001}
D.~Kuhn, How do people know?, Psychological science 12~(1) (2001) 1--8.
\newblock \href {http://dx.doi.org/10.1111/1467-9280.00302}
  {\path{doi:10.1111/1467-9280.00302}}.

\bibitem{Kulesza2013}
T.~Kulesza, S.~Stumpf, M.~Burnett, S.~Yang, I.~Kwan, W.-K. Wong, Too much, too
  little, or just right? {W}ays explanations impact end users' mental models,
  in: IEEE Symposium on Visual Languages and Human Centric Computing, IEEE,
  2013, pp. 3--10.
\newblock \href {http://dx.doi.org/10.1109/VLHCC.2013.6645235}
  {\path{doi:10.1109/VLHCC.2013.6645235}}.

\bibitem{Tullio2007}
J.~Tullio, A.~K. Dey, J.~Chalecki, J.~Fogarty, How it works: A field study of
  non-technical users interacting with an intelligent system, in: Proceedings
  of the 2007 Conference on Human Factors in Computing Systems (CHI),
  Association for Computing Machinery, New York, NY, USA, 2007, pp. 31--40.
\newblock \href {http://dx.doi.org/10.1145/1240624.1240630}
  {\path{doi:10.1145/1240624.1240630}}.

\bibitem{Mitchell2019}
M.~Mitchell, S.~Wu, A.~Zaldivar, P.~Barnes, L.~Vasserman, B.~Hutchinson,
  E.~Spitzer, I.~D. Raji, T.~Gebru, Model cards for model reporting, in:
  Proceedings of the 2019 Conference on Fairness, Accountability, and
  Transparency, Association for Computing Machinery, New York, NY, USA, 2019,
  p. 220–229.
\newblock \href {http://dx.doi.org/10.1145/3287560.3287596}
  {\path{doi:10.1145/3287560.3287596}}.

\bibitem{Langer2018}
M.~Langer, C.~J. K{\"o}nig, A.~Fitili, Information as a double-edged sword: The
  role of computer experience and information on applicant reactions towards
  novel technologies for personnel selection, Computers in Human Behavior 81
  (2018) 19--30.
\newblock \href {http://dx.doi.org/10.1016/j.chb.2017.11.036}
  {\path{doi:10.1016/j.chb.2017.11.036}}.

\bibitem{newman2020}
D.~T. Newman, N.~J. Fast, D.~J. Harmon, When eliminating bias isn’t fair:
  Algorithmic reductionism and procedural justice in human resource decisions,
  Organizational Behavior and Human Decision Processes 160 (2020) 149--167.
\newblock \href {http://dx.doi.org/10.1016/j.obhdp.2020.03.008}
  {\path{doi:10.1016/j.obhdp.2020.03.008}}.

\bibitem{Bazire2005}
M.~Bazire, P.~Br{\'{e}}zillon, Understanding context before using it, in:
  A.~Dey, B.~Kokinov, D.~Leake, R.~Turner (Eds.), Modeling and Using Context,
  Springer, 2005, pp. 29--40.
\newblock \href {http://dx.doi.org/10.1007/11508373_3}
  {\path{doi:10.1007/11508373_3}}.

\bibitem{Dourish2004}
P.~Dourish, What we talk about when we talk about context, Personal and
  Ubiquitous Computing 8~(1) (2004) 19--30.
\newblock \href {http://dx.doi.org/10.1007/s00779-003-0253-8}
  {\path{doi:10.1007/s00779-003-0253-8}}.

\bibitem{Bobocel2005}
D.~R. Bobocel, A.~Zdaniuk, How can explanations be used to foster
  organizational justice, Handbook of organizational justice (2005) 469--498.

\bibitem{Folger2001}
R.~Folger, R.~Cropanzano, Fairness theory: Justice as accountability, Advances
  in organizational justice 1 (2001) 1--55.

\bibitem{Shaw2003}
J.~C. Shaw, E.~Wild, J.~A. Colquitt, To justify or excuse?: A meta-analytic
  review of the effects of explanations, Journal of Applied Psychology 88~(3)
  (2003) 444--458.
\newblock \href {http://dx.doi.org/10.1037/0021-9010.88.3.444}
  {\path{doi:10.1037/0021-9010.88.3.444}}.

\bibitem{Brockner1996}
J.~Brockner, B.~M. Wiesenfeld, An integrative framework for explaining
  reactions to decisions: Interactive effects of outcomes and procedures,
  Psychological Bulletin 120~(2) (1996) 189--208.
\newblock \href {http://dx.doi.org/10.1037/0033-2909.120.2.189}
  {\path{doi:10.1037/0033-2909.120.2.189}}.

\bibitem{Wang2020}
R.~Wang, F.~M. Harper, H.~Zhu, Factors influencing perceived fairness in
  algorithmic decision-making: Algorithm outcomes, development procedures, and
  individual differences, in: Proceedings of the 2020 Conference on Human
  Factors in Computing Systems (CHI), Association for Computing Machinery, New
  York, NY, USA, 2020, pp. 1--14.
\newblock \href {http://dx.doi.org/10.1145/3313831.3376813}
  {\path{doi:10.1145/3313831.3376813}}.

\bibitem{Lind2002}
E.~A. Lind, K.~van~den Bos, When fairness works: Toward a general theory of
  uncertainty management, Research in Organizational Behavior 24 (2002)
  181--223.
\newblock \href {http://dx.doi.org/10.1016/s0191-3085(02)24006-x}
  {\path{doi:10.1016/s0191-3085(02)24006-x}}.

\bibitem{Colquitt2002}
J.~A. Colquitt, J.~M. Chertkoff, Explaining injustice: The interactive effect
  of explanation and outcome on fairness perceptions and task motivation,
  Journal of Management 28~(5) (2002) 591--610.
\newblock \href {http://dx.doi.org/10.1177/014920630202800502}
  {\path{doi:10.1177/014920630202800502}}.

\bibitem{Liu2012}
P.~Liu, Z.~Li, Task complexity: A review and conceptualization framework,
  International Journal of Industrial Ergonomics 42~(6) (2012) 553--568.
\newblock \href {http://dx.doi.org/10.1016/j.ergon.2012.09.001}
  {\path{doi:10.1016/j.ergon.2012.09.001}}.

\bibitem{Wilkenfeld2014}
D.~A. Wilkenfeld, Functional explaining: A new approach to the philosophy of
  explanation, Synthese 191 (2014) 3367--3391.
\newblock \href {http://dx.doi.org/10.1007/s11229-014-0452-z}
  {\path{doi:10.1007/s11229-014-0452-z}}.

\bibitem{Wilkenfeld2016}
D.~A. Wilkenfeld, D.~Plunkett, T.~Lombrozo, Depth and deference: When and why
  we attribute understanding, Philosophical Studies 173 (2016) 373--393.
\newblock \href {http://dx.doi.org/10.1007/s11098-015-0497-y}
  {\path{doi:10.1007/s11098-015-0497-y}}.

\bibitem{Lombrozo2011}
T.~Lombrozo, The instrumental value of explanations, Philosophy Compass 6~(8)
  (2011) 539--551.
\newblock \href {http://dx.doi.org/10.1111/j.1747-9991.2011.00413.x}
  {\path{doi:10.1111/j.1747-9991.2011.00413.x}}.

\bibitem{Williams2013}
J.~J. Williams, T.~Lombrozo, Explanation and prior knowledge interact to guide
  learning, Cognitive Psychology 66~(1) (2013) 55--84.
\newblock \href {http://dx.doi.org/10.1016/j.cogpsych.2012.09.002}
  {\path{doi:10.1016/j.cogpsych.2012.09.002}}.

\bibitem{Lombrozo2012}
T.~Lombrozo, B.~Rehder, Functions in biological kind classification, Cognitive
  Psychology 65~(4) (2012) 457--485.
\newblock \href {http://dx.doi.org/10.1016/j.cogpsych.2012.06.002}
  {\path{doi:10.1016/j.cogpsych.2012.06.002}}.

\bibitem{Hempel1965}
C.~G. Hempel, Deductive-nomological explanation, in: Aspects of Scientific
  Explanation, Free Press, 1965, pp. 335--376.

\bibitem{Salmon1984}
W.~C. Salmon, Scientific explanation and the causal structure of the world,
  Princeton University Press, 1984.

\bibitem{Gardenfors1988}
P.~G{\"a}rdenfors, Knowledge in flux: Modeling the dynamics of epistemic
  states, MIT Press, 1988.

\bibitem{Wachter2018}
S.~Wachter, B.~Mittelstadt, C.~Russell, Counterfactual explanations without
  opening the black box: Automated decisions and the {GDPR}, Harv. JL \& Tech.
  31~(2).
\newblock \href {http://dx.doi.org/10.2139/ssrn.3063289}
  {\path{doi:10.2139/ssrn.3063289}}.

\bibitem{Craver2007}
C.~F. Craver, Explaining the Brain, Oxford University Press, Oxford, 2007.
\newblock \href {http://dx.doi.org/10.1093/acprof:oso/9780199299317.001.0001}
  {\path{doi:10.1093/acprof:oso/9780199299317.001.0001}}.

\bibitem{Pearl2009}
J.~Pearl, Causality: Models, reasoning, and inference, Cambridge University
  Press, 2009, 2nd edition.
\newblock \href {http://dx.doi.org/10.1017/CBO9780511803161}
  {\path{doi:10.1017/CBO9780511803161}}.

\bibitem{Spirtes2001}
P.~Spirtes, C.~Glymour, R.~Scheines, Causation, Prediction and Search, MIT
  Press, 2001, 2nd edition.
\newblock \href {http://dx.doi.org/10.7551/mitpress/1754.001.0001}
  {\path{doi:10.7551/mitpress/1754.001.0001}}.

\bibitem{Borsboom2018}
D.~Borsboom, A.~Cramer, A.~Kalis, Brain disorders? {N}ot really… {W}hy
  network structures block reductionism in psychopathology research, Behavioral
  and Brain Sciences 42 (2018) 1--54.
\newblock \href {http://dx.doi.org/10.1017/S0140525X17002266}
  {\path{doi:10.1017/S0140525X17002266}}.

\bibitem{Lombrozo2007}
T.~Lombrozo, Simplicity and probability in causal explanation, Cognitive
  Psychology 55~(3) (2007) 232--257.
\newblock \href {http://dx.doi.org/10.1016/j.cogpsych.2006.09.006}
  {\path{doi:10.1016/j.cogpsych.2006.09.006}}.

\bibitem{Vasilyeva2017}
N.~Vasilyeva, D.~Wilkenfeld, T.~Lombrozo, Contextual utility affects the
  perceived quality of explanations, Psychonomic Bulletin \& Review 24 (2017)
  1436--1450.
\newblock \href {http://dx.doi.org/10.3758/s13423-017-1275-y}
  {\path{doi:10.3758/s13423-017-1275-y}}.

\bibitem{Bellotti2001}
V.~Bellotti, K.~Edwards, Intelligibility and accountability: Human
  considerations in context-aware systems, Human-Computer Interaction 16~(2-4)
  (2001) 193--212.
\newblock \href {http://dx.doi.org/10.1207/s15327051hci16234_05}
  {\path{doi:10.1207/s15327051hci16234_05}}.

\bibitem{Hartley2001}
K.~Hartley, L.~D. Bendixen, Educational research in the internet age: Examining
  the role of individual characteristics, Educational researcher 30~(9) (2001)
  22--26.
\newblock \href {http://dx.doi.org/10.3102/0013189X030009022}
  {\path{doi:10.3102/0013189X030009022}}.

\bibitem{Kauffman2015}
H.~Kauffman, A review of predictive factors of student success in and
  satisfaction with online learning, Research in Learning Technology 23.
\newblock \href {http://dx.doi.org/10.3402/rlt.v23.26507}
  {\path{doi:10.3402/rlt.v23.26507}}.

\bibitem{McNamara1996}
D.~S. McNamara, E.~Kintsch, N.~B. Songer, W.~Kintsch, Are good texts always
  better? interactions of text coherence, background knowledge, and levels of
  understanding in learning from text, Cognition and Instruction 14~(1) (1996)
  1--43.
\newblock \href {http://dx.doi.org/10.1207/s1532690xci1401\_1}
  {\path{doi:10.1207/s1532690xci1401\_1}}.

\bibitem{Goldberg1981}
L.~Goldberg, Language and individual differences: The search for universals in
  personality lexicons, in: W.~L. (Ed.), Review of Personality and Social
  Psychology, vol. 2 Edition, {SAGE} Publications, 1981, pp. 141--166.

\bibitem{Cacioppo1982}
J.~T. Cacioppo, R.~E. Petty, The need for cognition, Journal of Personality and
  Social Psychology 42~(1) (1982) 116--131.
\newblock \href {http://dx.doi.org/10.1037/0022-3514.42.1.116}
  {\path{doi:10.1037/0022-3514.42.1.116}}.

\bibitem{Haugtvedt1992}
C.~P. Haugtvedt, R.~E. Petty, Personality and persuasion: Need for cognition
  moderates the persistence and resistance of attitude changes, Journal of
  Personality and Social Psychology 63~(2) (1992) 308--319.
\newblock \href {http://dx.doi.org/10.1037/0022-3514.63.2.308}
  {\path{doi:10.1037/0022-3514.63.2.308}}.

\bibitem{DeBacker2009}
T.~K. DeBacker, H.~M. Crowson, The influence of need for closure on learning
  and teaching, Educational Psychology Review 21 (2009) 303--323.
\newblock \href {http://dx.doi.org/10.1007/s10648-009-9111-1}
  {\path{doi:10.1007/s10648-009-9111-1}}.

\bibitem{Webster1994}
D.~M. Webster, A.~W. Kruglanski, Individual differences in need for cognitive
  closure, Journal of Personality and Social Psychology 67~(6) (1994)
  1049--1062.
\newblock \href {http://dx.doi.org/10.1037/0022-3514.67.6.1049}
  {\path{doi:10.1037/0022-3514.67.6.1049}}.

\bibitem{Fernbach2013}
P.~M. Fernbach, S.~A. Sloman, R.~S. Louis, J.~N. Shube, Explanation fiends and
  foes: How mechanistic detail determines understanding and preference, Journal
  of Consumer Research 39~(5) (2012) 1115--1131.
\newblock \href {http://dx.doi.org/10.1086/667782} {\path{doi:10.1086/667782}}.

\bibitem{Hasher1988}
L.~Hasher, R.~T. Zacks, Working memory, comprehension, and aging: A review and
  a new view, in: Psychology of Learning and Motivation, Elsevier, 1988, pp.
  193--225.
\newblock \href {http://dx.doi.org/10.1016/s0079-7421(08)60041-9}
  {\path{doi:10.1016/s0079-7421(08)60041-9}}.

\bibitem{Ackerman2012}
R.~Ackerman, T.~Lauterman, Taking reading comprehension exams on screen or on
  paper? {A} metacognitive analysis of learning texts under time pressure,
  Computers in Human Behavior 28~(5) (2012) 1816--1828.
\newblock \href {http://dx.doi.org/10.1016/j.chb.2012.04.023}
  {\path{doi:10.1016/j.chb.2012.04.023}}.

\bibitem{Prewett2010}
M.~S. Prewett, R.~C. Johnson, K.~N. Saboe, L.~R. Elliott, M.~D. Coovert,
  Managing workload in human{\textendash}robot interaction: A review of
  empirical studies, Computers in Human Behavior 26~(5) (2010) 840--856.
\newblock \href {http://dx.doi.org/10.1016/j.chb.2010.03.010}
  {\path{doi:10.1016/j.chb.2010.03.010}}.

\bibitem{Starcke2008}
K.~Starcke, O.~T. Wolf, H.~J. Markowitsch, M.~Brand, Anticipatory stress
  influences decision making under explicit risk conditions, Behavioral
  Neuroscience 122~(6) (2008) 1352--1360.
\newblock \href {http://dx.doi.org/10.1037/a0013281}
  {\path{doi:10.1037/a0013281}}.

\bibitem{Lupien2007}
S.~Lupien, F.~Maheu, M.~Tu, A.~Fiocco, T.~Schramek, The effects of stress and
  stress hormones on human cognition: Implications for the field of brain and
  cognition, Brain and Cognition 65~(3) (2007) 209--237.
\newblock \href {http://dx.doi.org/10.1016/j.bandc.2007.02.007}
  {\path{doi:10.1016/j.bandc.2007.02.007}}.

\bibitem{Hancock2007}
P.~A. Hancock, On the process of automation transition in multitask
  human{\textendash}machine systems, {IEEE} Transactions on Systems, Man, and
  Cybernetics - Part A: Systems and Humans 37~(4) (2007) 586--598.
\newblock \href {http://dx.doi.org/10.1109/tsmca.2007.897610}
  {\path{doi:10.1109/tsmca.2007.897610}}.

\bibitem{Chazette2019}
L.~{Chazette}, O.~{Karras}, K.~{Schneider}, Do end-users want explanations?
  {A}nalyzing the role of explainability as an emerging aspect of
  non-functional requirements, in: IEEE 27th International Requirements
  Engineering Conference (RE), 2019, pp. 223--233.
\newblock \href {http://dx.doi.org/10.1109/RE.2019.00032}
  {\path{doi:10.1109/RE.2019.00032}}.

\bibitem{Chazette2020}
L.~Chazette, K.~Schneider, Explainability as a non-functional requirement:
  challenges and recommendations, Requirements Engineering 25~(4) (2020)
  493--514.
\newblock \href {http://dx.doi.org/10.1007/s00766-020-00333-1}
  {\path{doi:10.1007/s00766-020-00333-1}}.

\bibitem{Arya2019}
V.~Arya, R.~K.~E. Bellamy, P.-Y. Chen, A.~Dhurandhar, M.~Hind, S.~C. Hoffman,
  S.~Houde, Q.~V. Liao, R.~Luss, A.~Mojsilovi{\'c}, S.~Mourad, P.~Pedemonte,
  R.~Raghavendra, J.~T. Richards, P.~Sattigeri, K.~Shanmugam, M.~Singh, K.~R.
  Varshney, D.~Wei, Y.~Zhang, One explanation does not fit all: {A} toolkit and
  taxonomy of {AI} explainability techniques, CoRR abs/1909.03012.
\newblock \href {http://arxiv.org/abs/1909.03012} {\path{arXiv:1909.03012}}.

\bibitem{Woodward2019}
J.~Woodward, Scientific explanation, in: E.~N. Zalta (Ed.), The Stanford
  Encyclopedia of Philosophy, winter 2019 Edition, Metaphysics Research Lab,
  Stanford University, 2019, pp. 1--101.

\bibitem{Hall2019}
M.~Hall, D.~Harborne, R.~Tomsett, V.~Galetic, S.~Quintana-Amate, A.~Nottle,
  A.~Preece, A systematic method to understand requirements for explainable {AI
  (XAI)} systems, in: Proceedings of the IJCAI 2019 Workshop on Explainable
  Artificial Intelligence (XAI), 2019, pp. 21--27.

\bibitem{Miller2017}
T.~Miller, P.~Howe, L.~Sonenberg, Explainable {AI}: Beware of inmates running
  the asylum. or: How {I} learnt to stop worrying and love the social and
  behavioural sciences, in: Proceedings of the IJCAI 2017 Workshop on
  Explainable Artificial Intelligence (XAI), 2017, pp. 36--42.

\bibitem{Kim2014}
B.~Kim, C.~Rudin, J.~A. Shah, The bayesian case model: A generative approach
  for case-based reasoning and prototype classification, in: Advances in Neural
  Information Processing Systems, 2014, pp. 1952--1960.

\bibitem{Kim2016}
B.~Kim, R.~Khanna, S.~Koyejo, Examples are not enough, learn to criticize!
  {C}riticism for interpretability, in: Advances in Neural Information
  Processing Systems, 2016, pp. 2280--2288.

\bibitem{Carroll1987}
J.~M. Carroll, M.~B. Rosson, Paradox of the active user, in: Interfacing
  Thought: Cognitive Aspects of Human-Computer Interaction, MIT Press,
  Cambridge, MA, USA, 1987, pp. 80--111.

\end{thebibliography}

\end{document}